\documentclass{article}

\usepackage[preprint]{neurips_2023}
\usepackage{listings}
\usepackage[utf8]{inputenc}
\usepackage{textgreek}
\usepackage{natbib}
\usepackage{listings}
\usepackage{xcolor}
\usepackage{makecell}

\usepackage[most]{tcolorbox}

\newtcolorbox{promptbox}{
    colback=gray!5,
    colframe=gray!50,
    boxrule=0.5pt,
    arc=2pt,
    left=6pt,
    right=6pt,
    top=6pt,
    bottom=6pt,
    fontupper=\small\ttfamily,
    breakable
}

\usepackage{tcolorbox}
\tcbuselibrary{breakable,skins,listings}
\tcbset{
  scopebase/.style={
    enhanced, breakable,
    colback=gray!4, colframe=black!65, colbacktitle=black!65, coltitle=white,
    boxrule=0.5pt, arc=1.5pt,
    left=5pt, right=5pt, top=4pt, bottom=4pt,
    fonttitle=\bfseries\small, fontupper=\small,
  }
}
\newtcolorbox{artifactbox}[1]{scopebase, title={#1}}
\newtcblisting{codebox}[1]{scopebase, title={#1}, listing only,
  listing options={style=pythonstyle}}

\tcbset{
  scopeinner/.style={
    enhanced,
    colback=white, colframe=black!35, colbacktitle=gray!18, coltitle=black,
    boxrule=0.4pt, arc=1pt,
    left=4pt, right=4pt, top=3pt, bottom=3pt,
    fonttitle=\bfseries\footnotesize, fontupper=\small,
    before skip=4pt, after skip=4pt,
  }
}
\newtcolorbox{innerbox}[1]{scopeinner, title={#1}}
\newtcblisting{innercode}[1]{scopeinner, title={#1}, listing only,
  listing options={style=pythonstyle, basicstyle=\ttfamily\scriptsize}}

\lstdefinestyle{xmlsample}{
  basicstyle=\ttfamily\footnotesize,
  columns=fullflexible, keepspaces=false,
  breaklines=true, breakatwhitespace=false,
  breakindent=1em, postbreak={},
  alsoletter={.,;:!?</>"},
  frame=none, xleftmargin=0pt, aboveskip=2pt, belowskip=2pt,
}

\lstdefinestyle{prompt}{
    basicstyle=\scriptsize\linespread{1.25}\selectfont,
    columns=fullflexible,
    keepspaces=true,
    showstringspaces=false,
    breaklines=true,
    breakatwhitespace=false,
    breakindent=0pt,
    postbreak={},
    alsoletter={.,;:!?</>"},
    frame=none, xleftmargin=0pt, aboveskip=2pt, belowskip=2pt,
    literate=
        {→}{{$\rightarrow$}}1
        {—}{{\textemdash}}1
        {–}{{\textendash}}1
        {…}{{\ldots}}1
        {“}{{\textquotedblleft}}1
        {”}{{\textquotedblright}}1
        {‘}{{\textquoteleft}}1
        {’}{{\textquoteright}}1
}

\lstdefinelanguage{JSON}{
    basicstyle=\ttfamily\small,
    breaklines=true,
    frame=none,
    numbers=none,
    numberstyle=\tiny,
    showstringspaces=false,
    stringstyle=\color{red},
    morestring=[b]",
    literate=
     *{0}{{{\color{blue}0}}}{1}
      {1}{{{\color{blue}1}}}{1}
      {2}{{{\color{blue}2}}}{1}
      {3}{{{\color{blue}3}}}{1}
      {4}{{{\color{blue}4}}}{1}
      {5}{{{\color{blue}5}}}{1}
      {6}{{{\color{blue}6}}}{1}
      {7}{{{\color{blue}7}}}{1}
      {8}{{{\color{blue}8}}}{1}
      {9}{{{\color{blue}9}}}{1}
      {:}{{{\color{black}:}}}{1}
      {,}{{{\color{black},}}}{1}
      {\{}{{{\color{black}\{}}}{1}
      {\}}{{{\color{black}\}}}}{1}
      {[}{{{\color{black}[}}}{1}
      {]}{{{\color{black}]}}}{1},
}

\lstdefinestyle{pythonstyle}{
    language=Python,
    basicstyle=\ttfamily\footnotesize,
    keywordstyle=\color{blue},
    commentstyle=\color{gray},
    stringstyle=\color{green!50!black},
    numbers=left,
    numberstyle=\tiny\color{gray},
    numbersep=8pt,
    frame=single,
    breaklines=true,
    breakatwhitespace=false,
    showstringspaces=false,
    columns=fullflexible
}

\usepackage[utf8]{inputenc} 
\usepackage[T1]{fontenc}    

\usepackage{url}            
\usepackage{booktabs}       
\usepackage{amsfonts}       
\usepackage{nicefrac}       
\usepackage{soul}
\usepackage{microtype}      

\usepackage{amsmath}
\usepackage{CJKutf8}

\usepackage[table,dvipsnames]{xcolor}

\usepackage{tabstackengine}
\usepackage{nicematrix}
\usepackage{tabularx, booktabs}
\usepackage{tablefootnote}
\usepackage{amssymb}
\usepackage{minitoc}
\usepackage{tikz,graphics,color,fullpage,float,epsf,caption,subcaption}
\usepackage{graphicx}

\usepackage{algorithm}
\usepackage{algorithmic}

\usepackage{makecell}

\usepackage{tabulary,multirow,overpic}

\definecolor{citecolor}{RGB}{34,139,34}

\usepackage[breaklinks=true,letterpaper=true,colorlinks, citecolor=citecolor,bookmarks=false]{hyperref}

\newcolumntype{x}[1]{>{\centering\arraybackslash}p{#1pt}}

\newcommand{\app}{\raise.17ex\hbox{$\scriptstyle\sim$}}

\newlength\savewidth

\newcommand{\method}{\textsc{Scope}} 

\newcommand{\std}[1]{\,{\scriptsize$\pm$#1}}

\title{SCOPE: Compiling Natural-Language Policies into
Neurosymbolic Predicate Programs}
\author{Arnon Mazza\thanks{Equal contribution.} \\ Plurai \\ {\tt\small arnonm@plurai.ai}
  \And
  Elad Levi\footnotemark[1] \\ Plurai \\ {\tt\small eladl@plurai.ai}}
\begin{document}
\doparttoc 
\faketableofcontents 
\part{} 

\maketitle
\begin{abstract}
Policy enforcement for LLM agents has shifted from single-turn content moderation to compound, application-specific policies evaluated over multi-turn transcripts and tool-call trajectories. Fine-tuned guardrail models are efficient but degrade on compound policies, and while LLM-as-a-judge approaches achieve higher accuracy, they incur high inference cost and tend to produce unstable predictions. Neurosymbolic methods offer a natural alternative, decomposing the policy into narrow neural decisions composed under explicit logic. However, existing symbolic layers lack the required expressiveness and are optimized against fixed data rather than the policy itself. To address this, we introduce SCOPE, which compiles a policy into an executable program of neural predicates executed by small models, deterministic extractors, and code that composes their outputs. The program is optimized jointly with its training data, using adversarial generation of counterexamples that target where the program deviates from the policy. We evaluate SCOPE on a benchmark of real-world policy tasks against prompt optimization, task decomposition, and workflow generation methods, achieving the highest average accuracy across tasks and models.
Programs optimized with the weakest executor transfer to other models without further optimization. This allows them to be deployed in production on inexpensive models, where they reach accuracy comparable to frontier reasoning models.
Unlike these monolithic models, SCOPE produces substantially more stable predictions, and each verdict is interpretable through the predicate outputs that produced it. Additionally, we validate the key components of our method through an ablation study. SCOPE thus enables reliable, interpretable, and efficient policy enforcement at production scale.
\end{abstract}

\section{Introduction}
\label{sct:introduction}
Large language models are increasingly deployed to enforce the policies of the
organizations that run them, from content rules \cite{llamaguard,wildguard,shieldgemma}
and regulatory obligations \cite{datla2025executable} to application-specific constraints
on permissible system behavior \cite{xiang2025,mazza2026barred}. As these applications
advance from single-turn moderation to multi-turn assistants and tool-using
agents, both the policies and the inputs they govern grow more complex. The unit
of judgment becomes a long transcript \cite{hoover2025dynaguard,yao2024taubench}, a structured plan
\cite{hua2024trustagent}, or an agent trajectory \cite{xiang2025,shieldagent} rather than a single
message, and a modern policy is rarely one flat category: it bundles conditions,
exceptions, counts, and interactions that must be assessed together
\cite{hoover2025dynaguard}. Adhering to such a policy is therefore both critical and
hard \cite{yao2024taubench}, as the verdict turns on a compound, context-dependent specification rather
than a single surface feature, where one overlooked clause produces a false
violation or a missed obligation.

The most direct approach uses a language model as the policy classifier itself.
Its principal advantage is generality: any policy that can be expressed in
natural language can be enforced without a predefined taxonomy
\cite{hoover2025dynaguard,palla2025policy,ossafeguard}. This approach takes two forms. The first trains a custom
classifier on the task, typically a small fine-tuned guardrail model
\cite{llamaguard,wildguard,shieldgemma,aegis}. Such models are fast and
inexpensive, and on a single, well-scoped rule they match far larger models
\cite{mazza2026barred}. Their capacity is the ceiling, so accuracy falls as a policy
compounds interacting conditions beyond what one model resolves in a
single pass \cite{dziri2023}. The second prompts a frontier model with the
policy at runtime and reads its verdict, the paradigm known as LLM-as-a-judge
\cite{mtbench,liu2023geval}. While it is more capable on the compound case, its predictions are unstable \cite{haldar2025rating,schroeder2024trust,atil2024nondeterminism}, with variance rising with the number of clauses in the
input \cite{gsmsymbolic}. Further, even dedicated reasoning models collapse past a complexity threshold rather than scaling through it \cite{illusion}. Finally, its inference cost and latency make it impractical at production scale, where high-throughput classification is required. The first form is thus reliable only on simple policies, while the second is neither stable nor scalable.

\begin{figure}[t]
\centering
\includegraphics[width=\textwidth]{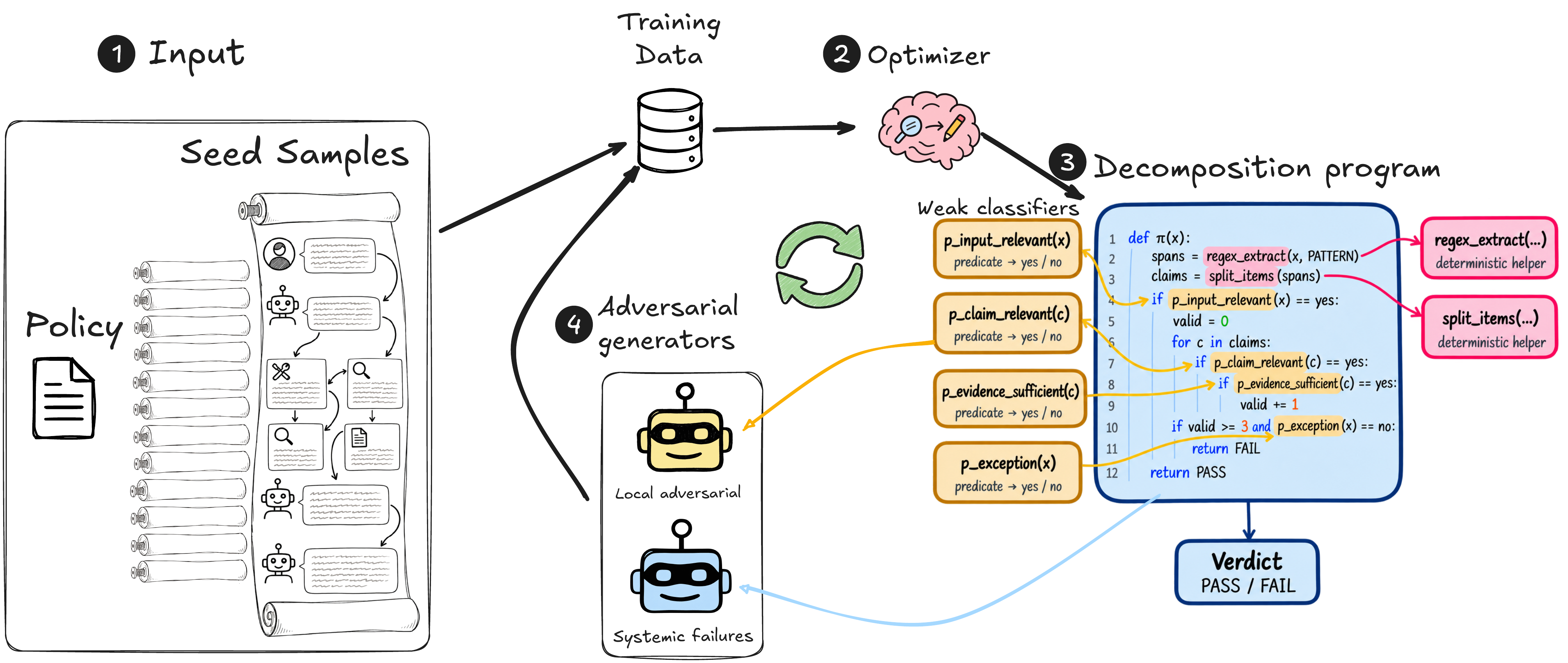}
\caption{(1) Given a natural-language policy and a small pool of seed samples, (2) the optimizer compiles the policy into (3) a decomposition program: predicates answered by weak models, deterministic extractors, and Python code that composes their outputs into a verdict. (4) Adversarial generators extend the training pool with counterexamples where the program departs from the policy, at the level of a single predicate (local) or the composition as a whole (systemic). The loop repeats, optimizing the program and its training data jointly.}
\label{fig:overview}
\end{figure}

A neurosymbolic approach is the natural alternative. Such methods pair neural
components that supply semantic judgment with a symbolic layer that composes
their outputs under explicit logic \cite{deepproblog,scallop}. Because each sub-decision is narrowly scoped and the composition is explicit, the verdict is more stable than a monolithic judgment \cite{lee2025checkeval}. As narrow sub-decisions are simpler \cite{khot2023decomposed,zhou2023leasttomost,adapt}, they can also be answered by small models, which makes the system cheaper to run. The weakness of existing methods lies in the symbolic layer itself, along two axes. The first is expressiveness: checklists \cite{lee2025checkeval,bineval}, weighted rule sets \cite{yang2025rlie,rbr}, and compiled Boolean or probabilistic rules \cite{shieldagent,r2guard,quadsentinel} cannot express the conditional, iterative, and computational structure that complex real-world use cases carry. The second is bias: a learned or synthesized layer is fit to a proxy for the policy, not to the policy itself, whether a rule set tuned to its training samples \cite{yang2025rlie}, a program searched against a static benchmark \cite{zhang2025aflow,hu2025automated}, or a function verified against self-generated tests and specifications \cite{miculicich2025veriguard}.

In this paper, we introduce SCOPE (Symbolic Composition Over Predicates and Extractors), which compiles a natural-language policy into a neurosymbolic program: neural predicates resolved by weak models, cheap sub-frontier models served without extended reasoning, deterministic extractors for exact computation, and executable code that composes their outputs. Expressiveness is met by the composition being code: it expresses the structure of real-world policies naturally, which rule-based formalisms capture only at low fidelity. Bias is mitigated by how the training data is built. Rather than relying only on a static training set, the search draws on an adversary that repeatedly generates counterexamples targeting where the current program departs from the specification \cite{cegis}, and the program is revised against them.
Counterexamples are produced by the adversary at two levels: local drift, where a predicate no longer tracks the clause it transcribes, and global drift, where the decomposition as a whole misses a predicate or misroutes an interaction between clauses. Each counterexample is labeled by a strong teacher applying the full policy, and the cost is paid once at compilation. The unit of optimization is thus not a single rule \cite{mazza2026barred} but the compound policy: each round adds counterexamples and updates both the predicate formulations and the code that composes them. The converged program runs entirely on weak models, so it is cheap to serve. Each verdict is attributable to the predicate outputs behind it rather than a post-hoc rationale, and, being built from narrow classifications, is more stable than a monolithic judgment.

We release a benchmark of four policy-enforcement tasks, derived from public benchmarks or modeled on deployed agentic systems. The tasks span long agent traces, multi-turn counting, critic auditing, and dynamic multi-label outputs. SCOPE attains the best accuracy in most task-executor settings, outperforming prompt optimization, neurosymbolic methods, and workflow generation. Its advantage grows as the executor weakens. Once a policy is compiled into predicates a weak model can answer without reasoning, accuracy becomes largely insensitive to the choice of executor. Programs compiled with our weakest executor transfer directly to other models, matching frontier reasoning models at a fraction of the inference cost. They are also far more stable: where reasoning models flip a substantial share of their verdicts across repeated runs, SCOPE’s predictions remain nearly fixed, and each verdict traces to the predicate outputs that produced it. Ablations confirm that adversarial generation is central to these gains. More broadly, SCOPE shows that natural-language policies can be compiled rather than interpreted, turning policy enforcement from an opaque judgment delegated to a frontier model into a program that can be optimized, inspected, and deployed at scale.

\section{Related Work}

\paragraph{Policy enforcement.}
Policy enforcement asks a classifier to decide whether an input satisfies a policy written in prose. Modern deployments impose requirements that earlier content moderation did not. Policies may involve complex combinations of conditions, exceptions, counts, and interactions that must be assessed together, potentially over long inputs such as multi-turn transcripts \citep{hoover2025dynaguard,yao2024taubench}, structured plans \citep{hua2024trustagent}, or agent trajectories \citep{xiang2025,shieldagent}, in which the evidence deciding the verdict occupies a small fraction of the context. On such a
policy the verdict must nonetheless be accurate, since one overlooked clause
yields a false violation or a missed obligation. It must also be
cheap, because the enforcer gates every interaction and its cost is paid on each
one \citep{mazza2026barred,zheng2024lightweight}, and consistent, because a verdict that
shifts with position or surface form is not an enforcement of the policy at all
\citep{unfairevaluators,feng2025rightway,ye2025justice}.

Existing approaches trade expressiveness against cost and consistency. Blocklists and
regular expressions are the least expressive: deterministic, perfectly
repeatable, and nearly free to run, but blind to anything their patterns do not
literally contain \citep{spertus,nobata}. Compact models fine-tuned on a single
narrow task judge semantics rather than surface form
\citep{llamaguard,wildguard,shieldgemma,aegis,guardreasoner}, and need no annotation once their
training data is synthesized from the policy text alone
\citep{mazza2026barred,constclassifiers,chua2024flexible}; their capacity is the ceiling,
and accuracy falls once a policy bundles interacting conditions that a single forward pass must resolve together \citep{dziri2023,gsmsymbolic}. Guardrails conditioned on a policy supplied at runtime
\citep{hoover2025dynaguard,padhi2024granite,glider,prometheus} are expressive and cheap, but, being far smaller than frontier models, they share the capacity ceiling above and still err on multi-clause policies \citep{hoover2025dynaguard}. Frontier LLM-as-a-judge \citep{mtbench,liu2023geval} is the most expressive and is accurate on compound policies. Yet it is the least consistent \citep{haldar2025rating,schroeder2024trust}, variance rising with the number of clauses \citep{gsmsymbolic}, and the most expensive, a frontier call for every gated interaction. Prompt optimization \citep{ape,pryzant2023protegi,opro,wang2024promptagent,dspy,mipro,textgrad,cheng2024trace,agrawal2026gepa,ipc} can tune the judge's instructions, but as the verdict of a monolithic judge still comes from a single call, it inherits the same limitations.

\method{} breaks this trade-off. A policy is compiled into a program whose atoms
are predicates resolved by small models and whose composition is explicit code,
so the expressiveness of a frontier judge is obtained at the cost and the
repeatability of a weak classifier: no strong-model call is made at inference, the
composition is deterministic, and both the predicates and the code composing them
are optimized against the policy itself.

\paragraph{Neurosymbolic programs.}
Neurosymbolic methods pair neural components, which supply semantic judgment,
with a symbolic layer that composes their outputs. Keeping the composition
explicit rather than folding it into a network's weights makes the verdict
readable and checkable, and captures procedural structure that a single forward
pass does not \citep{nesyprog}. Two axes organize what follows: how expressive the symbolic layer is, and how much of the method is learned rather than fixed in advance. On the first axis, the layer ranges from a logic program \citep{deepproblog,scallop} to executable code \citep{gao2023pal,chen2023pot,li2024chaincode,visprog,vipergpt,codevqa,binder}, which, unlike a logic program whose rules are written by hand and whose neural predicates are trained through its inference, can be written and revised by an LLM and can call models directly on intermediate results.
On the second axis, both the composition and the neural predicates it calls may be either given or learned.
Classical differentiable programs fix the structure and train the predicates through it \citep{deepproblog,scallop,nmn}, while others search for the structure as well: over a typed functional language \citep{houdini}, guided by neural relaxations \citep{near}, or over programs whose nodes invoke a language model \citep{zhang2025aflow,hu2025automated,wang2025interpretable}.

Existing methods have stayed on one side of each axis at a time. One line
derives a symbolic structure, generated from the policy or fit to data, but over a restricted formalism: a
count over atomic yes/no questions \citep{lee2025checkeval,bineval}, linear weights over
natural-language rules \citep{rbr,yang2025rlie}, a probabilistic circuit over trajectory
predicates \citep{shieldagent}, or a logical encoding checked by an SMT solver \citep{an2025neurosymbolic}.
Such a formalism is tractable to derive precisely because it is restricted, and for the same reason it cannot make which question is asked depend on earlier answers, count occurrences over spans extracted from the input, or repeat a check over parts of the input selected at run time, structures that real policies commonly contain.
The other line
writes the composition as code, gaining that expressiveness, but leaves the
program fixed: hand-authored rails \citep{nemoguardrails}, a synthesized function
verified against self-generated tests and specifications \citep{miculicich2025veriguard,toolguard}, or a program emitted
per task whose generator is never tuned against the policy
\citep{pajama,xiang2025}. Its predicates are frozen prompts, so one
that drifts from the clause it transcribes stays drifted. SCOPE combines both lines: its composition is ordinary Python, and its structure and predicates are optimized jointly against the policy.

\paragraph{Curriculum learning.}
A training pool synthesized once from the task description
\citep{ye2022zerogen,sagert,mazza2026barred} fixes the distribution
before optimization begins. Curriculum learning instead presents examples in a
deliberate order, from easy to hard rather than at random \citep{curriculum}. The
order need not be set in advance: self-paced
learning selects the next examples by the model's current competence
\citep{selfpaced}, and automated curricula place a teacher in the loop that picks
each round's examples to maximize the student's learning progress
\citep{autocurriculum,matiisen2017teacher,aclsurvey}. Across these, the schedule is regenerated as the
model improves, so that training mass tracks where the model is currently weak
rather than the base rate of the data.

One such curriculum builds its examples adversarially, generating rather than
ranking: instead of drawing from a fixed set, it produces the inputs on which the
current model fails \citep{ipc}. Human annotators write examples a target model gets wrong
\citep{anli,beattheai,dynabench}; red-teaming automates that search with a
generator \citep{redteamlm,curiosityrt,rainbowteaming}; and in program synthesis,
counterexample-guided loops propose a candidate, obtain inputs on which a verifier
reports failure, and fold them back in \citep{cegis,sketching,sygus}.

\method{} folds this adversarial curriculum into the decomposition search, building the training data and optimizing the program together. Unlike the schemes above, its target is a program with named sub-components, so a counterexample points to the part at fault, and its verifier is a strong model reading the policy rather than a sound solver, which sacrifices formal guarantees for applicability to policies no solver accepts.

\section{Methods}
\label{sct:method}
 
SCOPE searches for an executable program $\pi$ that reproduces the verdicts of a natural-language policy when every predicate is answered by a weak model. The policy assigns each input, such as a conversation or an agent trace, a label from a fixed set or, for multi-label tasks, a set of labels. As detailed in Section~\ref{sec:program}, the program consists of \emph{predicates}, classification questions answered by a weak LLM, and \emph{extractors}, deterministic functions over the input, combined by Python composition code. We denote by $s(\pi, D)$ the score of $\pi$ on a pool $D$, measured as accuracy for single-label tasks and the Jaccard index for multi-label tasks.
Only a small seed pool $D_0$ of human-labeled samples is given. During optimization, an adversarial process generates new samples that target where the program may depart from the policy and adds them to the pool. The program is thus fit to the policy rather than to a fixed dataset.

Figure~\ref{fig:overview} illustrates the optimization process. It starts from a single predicate, the policy itself, and alternates two components. The optimizer executes $\pi$ on the current training pool $D_t$, uses the predicate traces to attribute each failure to a predicate or to the composition code, and revises the program. The adversarial generators then produce samples on which the revised program is likely to disagree with the policy, targeting both individual predicates and the composition as a whole. A teacher applying the full policy labels them to form $D_{t+1}$. The search stops at a threshold $\tau$ or after $K$ rounds, and returns the program that scores highest on $D_0$.
The search is carried out by strong LLM agents, whose prompts are given in Appendix~\ref{sct:app_prompts}. The predicates, by contrast, are part of the program and are answered by a weak model during the search. A program therefore scores well only if a weak model can execute it, matching the target deployment setting.

\subsection{The Decomposition Program}
\label{sec:program}

 \paragraph{Structure.}
A program consists of predicates, extractors, and composition code. A predicate is
a natural-language classification question with a fixed label set, typically
binary. An extractor is a deterministic function that returns a value, a span, or
a list of spans from the input. The composition code is a Python function that
invokes extractors to select spans, applies predicates to them, and maps the
resulting labels to a verdict. Predicates do not specify their input; the
composition code determines the text to which each predicate is applied. The same
predicate may therefore be applied to the full input, to a span within it, or
repeatedly within a single sample.

Expressing the composition as code allows the program to mirror the structure of the policy. A predicate can be restricted to the region its question concerns, such as one section of a long input, or applied multiple times to different parts of the input. Although predicate label sets are fixed, the program can handle tasks whose label set varies per sample or that require multiple labels, by determining how predicates are applied to the sample's candidates and how their outputs form the final prediction. Predicates can also be invoked conditionally on earlier answers. Counts and thresholds can be computed exactly by deterministic extractors, rather than estimated by a model \cite{gao2023pal,chen2023pot}.
Appendix~\ref{sct:app_task_decomposition_example} presents a complete program.
 
\paragraph{Execution.}
During the search, every predicate call is answered by an \emph{executor}, a sub-frontier model run with reasoning disabled and instructed to return only the label. The executor is chosen to match the capability of the models intended for deployment. A predicate that the executor cannot answer reliably is likely to be reworded or split into simpler questions. The structure of the program therefore adapts to the capability of the executor, and weaker executors yield programs with more predicates (Figure~\ref{fig:proxy_capability}b). Since the executor receives only a predicate and its input, without the policy, each predicate must be self-contained, stating the definitions, scope, and exceptions of the clause it encodes.

This choice of executor is motivated by cost, stability, and traceability. Sub-frontier models are considerably cheaper to run than frontier models, yet capable of answering narrowly scoped questions without any training. Running without reasoning is intended to keep predictions stable across runs (Section~\ref{sct:transfer}). Finally, executing a program on a sample produces a trace: the ordered sequence of predicate calls, each with the exact text it received and the label it returned. Traces make every verdict attributable to specific predicate outputs, and they are the primary signal for the search. The optimizer uses them to localize failures (Section~\ref{sec:optimizer}), and the local adversary uses them to observe the inputs a predicate actually receives (Section~\ref{sec:adversary}).
 
\subsection{Optimizer}
\label{sec:optimizer}

The search starts from the undecomposed task: a single predicate, the policy itself, applied to the full input. Before the first round, the teacher writes a rationale for each seed sample. At round $t$, the teacher labels the samples the adversaries added in the previous round (Section~\ref{sec:adversary}), and the current program is executed on $D_t$. The analyzer receives the policy, the program, and, for every sample, the predicted verdict, the gold label, its rationale, and the trace. Supervision exists only for the final verdict, so the analyzer infers which calls caused a failure by working backward: it replays the composition code on the recorded predicate labels, finds the labels that would have had to differ to produce the gold verdict, and inspects the text those calls received. Each failure is thus attributed to a predicate whose wording does not encode its clause, or to composition code that misencodes the policy. Correctly classified samples with similar rationales serve as a contrast set for checking candidate fixes against regressions.

The analyzer then recommends revisions: rewording or splitting a predicate, adding or removing a predicate or extractor, moving a condition between a predicate and an extractor, or rewriting the composition code. Each recommendation must be grounded in a clause, definition, or exception of the policy. A revision justified only by a phrase observed in a sample would fit the pool rather than the policy, and is excluded. The decomposer treats these recommendations as proposals and weighs them against the search history: previous programs and their scores, the revisions applied in the last round, and revisions already tried from the current program without improvement. When recent rounds do not improve the score, the search backtracks to the best program found so far, and the decomposer must pursue a substantially different direction, which prevents cycling among similar programs. The search terminates when $s(\pi, D_t)$ reaches a threshold $\tau$, or after $K$ rounds. Since $D_t$ grows across rounds, its scores are not comparable between rounds, so the returned program is the candidate with the highest score $s(\pi, D_0)$ on the fixed seed pool.
\subsection{Adversarial Generators}
\label{sec:adversary}

Optimizing against a fixed pool rewards fitting the pool: a policy condition that the pool rarely exercises can be omitted at little cost in accuracy, and the pool provides no signal to restore it. SCOPE therefore grows the pool adversarially. After each revision, two adversaries compare the policy with the current program and generate samples on which the two are likely to disagree. Supervision thus concentrates where the program is wrong, analogously to counterexample-guided synthesis~\cite{cegis}, with the policy as the specification. Both adversaries see the current pool, including $D_0$ and all previously generated samples, so new samples stay consistent with it and avoid redundancy.

The adversaries operate at two levels. The \emph{systemic} adversary targets global drift: a clause, exception, or interaction between clauses that the program as a whole misrepresents. It reads the full program and generates inputs on which the composition omits or misroutes part of the policy (Appendix~\ref{sct:app_adversarial_systemic}). The \emph{local} adversary targets local drift: a single predicate that no longer tracks the clause it encodes. It examines one predicate at a time, together with traces of the inputs that predicate receives, and generates inputs that probe where its wording diverges from the clause (Appendix~\ref{sct:app_adversarial_local}). Each sample is generated with an intended label and a justification. The teacher, applying the full policy, labels it without seeing either, and the sample enters the pool only if the two labels agree, so every admitted sample reflects two separate readings of the policy.

\begin{table}[t]
\centering
\small
\caption{Policy-enforcement benchmark of four tasks from real-world agentic systems. The tasks span input types, lengths, and label spaces, and each poses a distinct core challenge.}
\label{tab:datasets}
\setlength{\tabcolsep}{4.5pt}
\begin{tabular}{l l c l l c}
\addlinespace[0.5em]
\toprule
\textbf{Task} & \textbf{Input} & \textbf{Avg.\ length} & \textbf{Label space} & \textbf{Core challenge} & \textbf{Size} \\
\midrule
Coaching   & Message, critic output  & 260  & Pass / Fail / Bug    & Nested audit           & 174 \\
Escalation & Multi-turn dialogue     & 1.6k & True / False         & Cross-turn aggregation & 371 \\
Plan       & Full agent trace        & 45k  & Pass / Fail          & Sparse evidence        & 163 \\
Tile-match & Conversation, tile list & 630  & Dynamic, multi-label & Distractor candidates  & 163 \\
\bottomrule
\end{tabular}
\end{table}

\section{Experimental Setup}
\subsection{Datasets}
\label{sct:datasets}
We evaluate the decomposition framework on four tasks from different domains, described below. The datasets\footnote{All datasets are available at \url{https://huggingface.co/datasets/Plurai/SCOPE}} are summarized in Table~\ref{tab:datasets}, with additional details in Appendix~\ref{sct:app_dataset_details}.

\textbf{Coaching.}
In an agentic fitness and wellness system, one component generates short coaching messages, and an automated critic judges each message against a seven-dimension rubric. The task audits the critic: given a message with the critic's verdict and reasoning, output Pass, Fail, or Bug. A classifier must first judge the message itself, then check the critic's verdict and cited reasons against that judgment. Bug covers a wrong verdict, or a correct rejection with an inapplicable reason (Appendix~\ref{sct:app_adversarial_systemic}).

\textbf{Escalation.} Inspired by the counting policies of DynaGuard~\cite{hoover2025dynaguard}, the input is a customer-support dialogue. Escalation is warranted when the user re-raises the same unresolved intent at least three times, verbatim or rephrased. A turn that adds a new constraint, question, or information does not count, so each turn must be compared semantically with the earlier ones. The label is True when the agent's final reply escalates exactly when warranted (Appendix~\ref{sct:app_adversarial_local}).

\textbf{Plan.} Built on GAIA~\citep{mialon2024gaia} with authored samples, the input is a full agent trace ending in a research plan~\cite{smolagents}. A plan is valid if it references only available tools and collaborators, refers to tools abstractly, and ends with an \texttt{<end\_plan>} tag. A plan typically spans a few hundred characters within a trace of tens of thousands.

\begin{figure}[t]
\centering
\includegraphics[width=\textwidth]{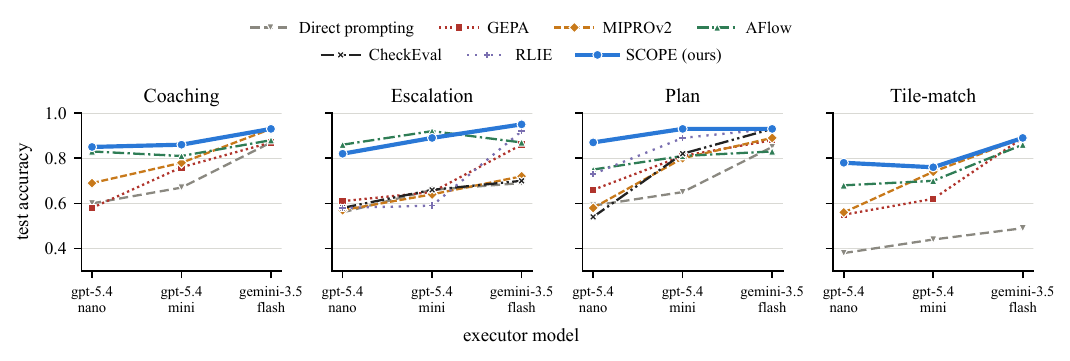}
\caption{Test accuracy by executor model and method, one panel per task. Values are from Table~\ref{tab:experiments}. Executors are ordered from weakest to strongest. CheckEval and RLIE are evaluated on Escalation and Plan only.
}
\label{fig:accuracy_plots}
\end{figure}

\textbf{Tile-match.} Given a conversation with a financial-dashboard assistant and the list of tiles on screen, the task identifies the tiles the user's last message refers to: one, several, or none. Each sample thus defines its own label space, and predictions are scored by the mean Jaccard index. Of the 124 test samples, 55 refer to no tile and serve as hard negatives against plausible distractors.

\subsection{Baseline methods}
\label{sct:baseline_models}
\textbf{Prompt optimization.}
To separate the effect of structural decomposition from prompt optimization alone, we compare against two optimizers implemented in DSPy~\cite{dspy,khattab2022demonstrate}: GEPA~\cite{agrawal2026gepa} and MIPROv2~\cite{mipro}. Both improve a fixed task description by rewriting it (MIPROv2 also selects few-shot demonstrations), without decomposing the task. GEPA maintains a pool of candidate instructions. It reflects on execution traces and metric feedback on a minibatch of training instances to propose a revision, and selects the next candidate to mutate from the Pareto front over training instances. MIPROv2 is a Bayesian optimizer that jointly searches over instructions and bootstrapped few-shot demonstrations, using a proposal model to draft candidate instructions grounded in a summary of the training data. For a controlled comparison, both optimize on the training pool that SCOPE starts from, use the same executor model, and are scored on the same held-out test set.

\textbf{Task decomposition.}
\label{sct:task_decomposition}
We evaluate \method{} with three executor models: gpt-5.4-nano, gpt-5.4-mini~\cite{gpt54}, and gemini-3.5-flash~\cite{gemini35flash}, each configured with reasoning disabled. We compare against two neurosymbolic task-decomposition baselines. 
CheckEval~\cite{lee2025checkeval} is a checklist-based LLM-as-a-judge that decomposes the policy into binary yes/no questions seeded from a manual list and expanded through LLM diversification and elaboration, and scores each sample by the proportion of ``yes'' answers.
RLIE~\cite{yang2025rlie} instead induces a set of natural-language rules from labeled samples, refined iteratively and combined by a logistic regression. By construction, both methods support only binary labels, so we evaluate them only on the binary tasks.

\textbf{Automated workflow generation.}
We compare against AFlow~\cite{zhang2025aflow}, an automated agentic-workflow optimizer, which runs Monte-Carlo tree search over programs that orchestrate multiple LLM calls through reusable operators. Like \method{} it searches over executable programs rather than a single prompt, but its search space is open-ended rather than a decomposition of the policy. Since \method{} and all other baselines were allowed to emit nothing but the label, we restricted AFlow to the Custom operator and disabled chain-of-thought answer generation. Notably, the instructions AFlow writes for the Custom operator may themselves elicit intermediate reasoning, on which the final label can be conditioned. This is inherent to the method, so we left it in place, thus allowing AFlow a capability that the other methods did not have.

\subsection{Implementation Details}
\label{sct:implementation_details}
All core models in the decomposition process (the analyzer, decomposer, adversarial sample generators and teacher) run Claude Opus 4.8~\cite{claudeopus48} in an agentic loop. The agent SDK~\cite{claudeagentsdk} provides file access, allowing the models to (i) read long records from disk on demand rather than receiving them inline, keeping prompts compact for tasks with inputs spanning tens of thousands of characters, and (ii) generate very long samples by editing seeds, which is difficult to accomplish in standard inference mode.
The executor models run the predicates, producing only a label, without a rationale.

For each task, we split the human-annotated set into training and test portions. The training portion forms the initial pool $D_0$. The test portion is used only for the final evaluation. The search runs for at most $K=8$ iterations, and stops early once accuracy on the current training pool reaches $\tau = 0.98$. In each iteration, the systemic adversary generates at most three samples and each predicate-local adversary adds at most one, so that adversarial samples enrich the human-labeled pool without dominating it.

\section{Results and Analysis}
\label{sct:results}

\begin{table}[t]
\vspace{-30pt}
\centering
\small
\caption{{Test accuracy with gpt-5.4-nano as the executor (mean Jaccard for Tile-match), averaged over three optimization runs, each evaluated three times, with standard deviations. \method{} outperforms all prompt-optimization and decomposition baselines on every task, and is the only method that remains strong across all four. Best in \textbf{bold}, second \protect\underline{underlined}.}}
\label{tab:main_nano}
\begin{tabular}{l c c c c}
\addlinespace[0.5em]
\toprule
\textbf{Method} & \textbf{Coaching} & \textbf{Escalation} & \textbf{Plan} & \textbf{Tile-match} \\
\midrule
Direct prompting & 0.60 \std{0.03} & 0.56 \std{0.03} & 0.59 \std{0.03} & 0.38 \std{0.03} \\
\midrule
\multicolumn{5}{l}{\textit{Prompt optimization}} \\
\addlinespace[0.2em]
GEPA         & 0.58 \std{0.01} & 0.61 \std{0.01} & 0.66 \std{0.02} & 0.55 \std{0.06} \\
MIPROv2      & 0.69 \std{0.01} & 0.57 \std{0.02} & 0.58 \std{0.03} & 0.56 \std{0.05} \\
\midrule
\multicolumn{5}{l}{\textit{Workflow generation}} \\
\addlinespace[0.2em]
AFlow        & \underline{0.83} \std{0.02} & \textbf{0.86} \std{0.03} & \underline{0.75} \std{0.08} & \underline{0.68} \std{0.09} \\
\midrule
\multicolumn{5}{l}{\textit{Task decomposition}} \\
\addlinespace[0.2em]
CheckEval    & N/A  & 0.58 \std{0.08} & 0.54 \std{0.04} & N/A  \\
RLIE         & N/A  & 0.58 \std{0.04} & 0.73 \std{0.05} & N/A  \\
SCOPE (ours) & \textbf{0.85} \std{0.03} & \underline{0.82} \std{0.02} & \textbf{0.87} \std{0.02} & \textbf{0.78} \std{0.03} \\
\bottomrule
\end{tabular}
\end{table}

\begin{figure*}[t]
\centering
\begin{subfigure}[b]{0.47\textwidth}
  \centering
  \includegraphics[width=\textwidth]{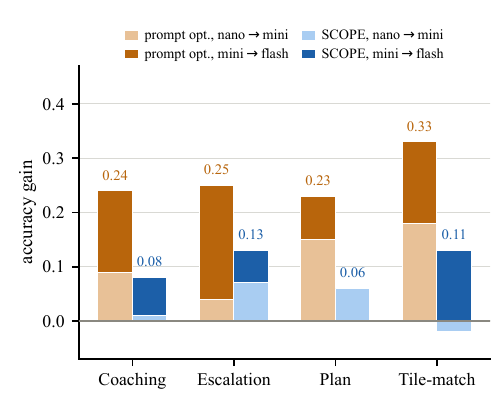}
  \caption{Accuracy gain from increasing executor strength.}
  \label{fig:decomp_gains_by_model}
\end{subfigure}
\hfill
\begin{subfigure}[b]{0.47\textwidth}
  \centering
  \includegraphics[width=\textwidth]{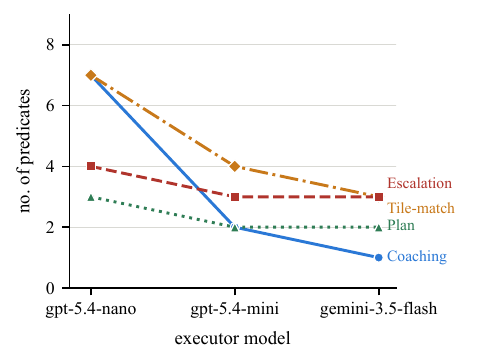}
  \caption{Number of predicates in the decomposition.}
  \label{fig:predicates}
\end{subfigure}
\caption{Effect of executor capability on the decomposition. (a) Accuracy gained when moving from the weakest to the strongest executor is substantially less for \method{} than for prompt optimization (average gain over tasks: 0.09 vs. 0.26). (b) Number of predicates in the final decomposition (median over repetitions). Executors are ordered from weakest to strongest by prompt-optimized accuracy, an ordering that holds on all four tasks.}
\label{fig:proxy_capability}
\end{figure*}

Figure~\ref{fig:accuracy_plots} reports test accuracy for the four tasks under the three executor models, with the exact numbers detailed in Table~\ref{tab:experiments} (Appendix~\ref{sct:app_detailed_results}). All reported numbers are averaged over three independent runs of the optimization process and three repetitions of the final test evaluation, for a total of 9 measurements. Both CheckEval and RLIE support binary predictions only and therefore apply to Escalation and Plan alone.

Comparing the executors on the undecomposed task after prompt optimization (maximum over GEPA and MIPROv2) reveals a consistent capability hierarchy: gpt-5.4-mini outperforms gpt-5.4-nano on all four tasks and gemini-3.5-flash outperforms gpt-5.4-mini on all four (mean accuracies over all tasks: 0.63, 0.75, 0.89).

\method{} attains the highest accuracy, or ties for it, in 10 of the 12 task-executor cells. AFlow follows, with the highest accuracy in the remaining 2, even though it has an inherent advantage of being the only method that can ask for reasoning in intermediate steps. \method{} is also consistent across runs: with gpt-5.4-nano as the executor, its standard deviation is at most 0.03 on every task, and a quarter to a third of AFlow's on Plan and Tile-match (Table~\ref{tab:main_nano}).

Figure~\ref{fig:accuracy_plots} shows that the gain \method{} achieves over direct prompting and prompt optimization widens as the executor weakens. This shows that the search finds decompositions that the least capable model can benefit from and execute with competitive performance. When a strong model is already able to handle the task, there is less room to improve through decompositions.

Further, Figure~\ref{fig:decomp_gains_by_model} shows that moving from the weakest executor to the strongest buys 0.26 accuracy on average under prompt optimization, but only 0.09 under \method{}. Decomposition thus compresses the capability hierarchy it was measured against: once the policy is expressed as predicates a small model can answer, accuracy becomes much less sensitive to the choice of executor.

The structure of the decomposition also responds to the executor: weaker models yield more predicates (Figure~\ref{fig:predicates}), with the largest increase from gpt-5.4-mini to gpt-5.4-nano. This reflects the design of the search: when the executor cannot answer a predicate accurately, the search further splits it.

Notably, \method{} reduces to prompt optimization when decomposition does not help: the search initializes with the undecomposed task and retains the best-scoring candidate, so a single predicate is always a reachable outcome. As an example, on the coaching task with gemini-3.5-flash, the search converged to a single predicate, with test accuracy increasing from 0.87 to 0.93.

\newcommand{\gain}[1]{{\scriptsize$\uparrow$#1}}
\newcommand{\loss}[1]{{\scriptsize$\downarrow$#1}}

\begin{table}[t]
\centering
\footnotesize
\setlength{\tabcolsep}{3pt}
\caption{Accuracy of the decompositions found with gpt-5.4-nano, executed by five open-weight models, without further optimization, and with reasoning disabled. Arrows show the gain over the same model on the undecomposed task. As a baseline, two high-end reasoning models run on the undecomposed task: gpt-5.5 with high reasoning effort, and gemini-3.1-pro with dynamic thinking.}
\label{tab:slms}
\begin{tabular}{l cccc}
\addlinespace[0.5em]
\toprule
\textbf{Executor} & \textbf{Coaching} & \textbf{Escalation} & \textbf{Plan} & \textbf{Tile-match} \\
\midrule
DeepSeek-V4.1-Flash (reasoning disabled) & 0.78 \gain{0.02} & \textbf{0.91} \gain{0.19} & \underline{0.86} \gain{0.33} & \textbf{0.89} \gain{0.34} \\
Qwen3-235B-A22B-Instruct-2507 & 0.90 \gain{0.27} & 0.78 \gain{0.04} & \underline{0.86} \loss{0.04} & \underline{0.86} \gain{0.37} \\
Qwen3-30B-A3B-Instruct-2507 & 0.87 \gain{0.04} & 0.78 \gain{0.15} & 0.83 \gain{0.12} & 0.81 \gain{0.37} \\
Gemma-3-27B-it              & 0.86 \gain{0.08} & 0.72 \gain{0.01} & 0.75 \gain{0.12} & 0.77 \gain{0.29} \\
Nemotron-3.5-Lightning      & 0.79 \gain{0.18} & 0.72 \gain{0.14} & 0.69 \gain{0.07} & 0.67 \gain{0.30} \\
\midrule
gpt-5.5 (high, undecomposed) & \underline{0.94} & \underline{0.89} & \textbf{0.89} & 0.82 \\
gemini-3.1-pro (dynamic, undecomposed) & \textbf{0.96} & 0.78 & \textbf{0.89} & 0.83 \\
\bottomrule
\end{tabular}
\vspace{-10pt}
\end{table}

\subsection{Cross-Executor Transferability}
\label{sct:transfer}
The simpler the predicates, the more likely a decomposition is to transfer to other executors. We examine this by running the decompositions found using our least capable executor, gpt-5.4-nano (whose evaluation was reported in Table~\ref{tab:main_nano}), on five open-weight models~\cite{deepseekv41flash,qwen3,qwen3instruct2507,gemma3,gemma3it,nemotron35lightning}, without any additional optimization. All five models run with reasoning disabled. Compared with each model's undecomposed run, the decompositions improve accuracy on average by roughly 0.11 on Coaching, Escalation and Plan, and 0.33 on Tile-match.
From a per-model perspective, the average gain ranges from 0.13 to 0.22 (Table~\ref{tab:slms}).

We compare the accuracy of these models running the decompositions with two high-end reasoning models evaluated directly on the undecomposed tasks: gpt-5.5~\cite{gpt55} with high reasoning effort, and gemini-3.1-pro~\cite{gemini31pro} with dynamic thinking. DeepSeek-V4.1-Flash and Qwen3-235B-A22B reach accuracy comparable to these models, at a fraction of the cost (Appendix~\ref{sct:app_cost_analysis}). DeepSeek attains the highest accuracy of all models on two tasks and the second-highest on a third, while Qwen3-235B-A22B ranks second on two tasks. Both are mixture-of-experts models that activate only a fraction of their parameters for each prediction (8B/16B during input processing/generation for DeepSeek~\cite{deepseekv41flash}, and 22B for Qwen~\cite{qwen3instruct2507}). The results are summarized in Table~\ref{tab:slms}.

Beyond accuracy and cost, decomposition provides a degree of stability and explainability that is absent from the monolithic high-end models. We evaluated prediction stability by running each model five times on the Tile-match dataset, whose multi-label structure requires multiple independent decisions per sample, making it particularly challenging from a stability perspective. DeepSeek-V4.1-Flash and Qwen3-235B-A22B were substantially more stable: only 5--7\% of samples changed their prediction at least once across the five runs, compared with 37\% for gpt-5.5 and 26\% for gemini-3.1-pro. As each prediction is traceable to a set of explicit predicates, this stability yields a more explainable and interpretable decision process than the opaque predictions of the monolithic models.

\subsection{Ablation: Removing Adversarial Samples}
To isolate the contribution of adversarial sample generation, we disabled both adversaries, so that the training pool remains fixed at $D_0$ throughout the search.

To achieve sufficient independent measurements for assessing statistical significance, for each setting we ran five independent optimization runs (for the Full variant, these include the three runs of Table~\ref{tab:main_nano}), and averaged three test evaluations per run output. Table~\ref{tab:ablation_adversary} compares the resulting accuracy with and without adversarial sample generation, using gpt-5.4-nano as the executor, with significance computed by a one-sided Welch's $t$-test comparing the five measurements of the two variants. Removing adversarial sample generation reduces accuracy on Coaching from 0.839 to 0.783 ($p=0.033$), on Plan from 0.865 to 0.813 ($p=0.005$), and on Tile-match from 0.782 to 0.744 ($p=0.045$). Adversarial sample generation thus improves accuracy on three of the four tasks, while having no effect on Escalation.

\begin{table}[t]
\vspace{-10pt}
\centering
\small
\caption{Ablation of adversarial generation with gpt-5.4-nano as the executor, averaged over five optimization runs. \textbf{Bold}: significant improvement ($p<0.05$, Welch's $t$-test).}
\label{tab:ablation_adversary}
\begin{tabular}{l c c c c}
\addlinespace[0.5em]
\toprule
\textbf{Variant} & \textbf{Coaching} & \textbf{Escalation} & \textbf{Plan} & \textbf{Tile-match} \\
\midrule
No adversary & 0.783 & 0.816 & 0.813 & 0.744 \\
Full         & \textbf{0.839} & 0.815 & \textbf{0.865} & \textbf{0.782} \\
\bottomrule
\end{tabular}
\vspace{-10pt}
\end{table}

\section{Conclusion}
We presented \method{}, which treats a natural-language policy as a specification to compile rather than a prompt to interpret, turning it into a program of neural predicates, deterministic extractors, and composition code. The program is optimized jointly with an adversarially grown training pool. Running the predicates with a weak executor steers the search toward predicates within its reach.
Across four policy-enforcement tasks, \method{} achieves the highest accuracy in most settings, with the largest gains under the weakest executor, and finds decompositions that are far less sensitive to the choice of executor.
Programs transfer to other executors without further optimization. When run by non-reasoning open-weight models, they reach accuracy comparable to frontier reasoning models, at a fraction of the cost, while producing substantially more stable verdicts, each traceable to the predicate outputs behind it.
Because the program is checked against a strong teacher rather than a formal verifier, its correctness rests on the teacher's reading of the policy, and verifying such programs remains an open problem. The underlying optimization, which co-evolves a program with the counterexamples that challenge it, is not specific to policy enforcement and may carry over to other tasks specified in natural language, such as rubric-based LLM evaluation and reward modeling. We release our benchmark to support further research on compound policy enforcement.

\bibliography{neurips_2023}

\newpage
\appendix
\part{Appendix} 

\section{Task decomposition example}
\label{sct:app_task_decomposition_example}
\begin{figure}[t]
\centering
\includegraphics[width=0.90\linewidth]{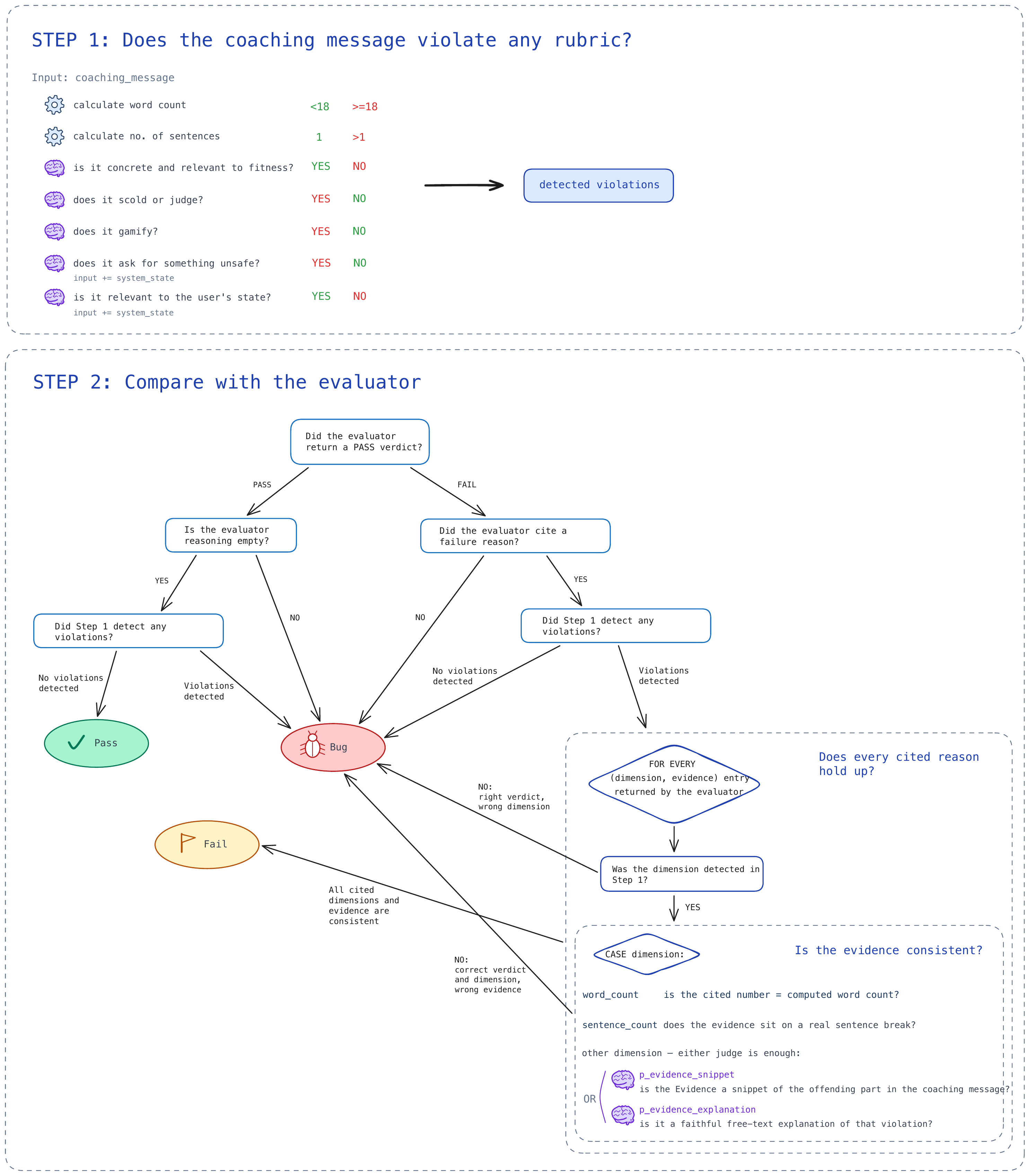}
\caption{Decomposition example for the Coaching task.}
\label{fig:fitness_decomp}
\end{figure}

Figure~\ref{fig:fitness_decomp} shows a possible decomposition of the Coaching task (Appendix~\ref{sct:app_adversarial_systemic}), obtained using gpt-5.4-nano as the executor. The input has four fields: the coaching message, the system state, and the evaluator's verdict and reasoning. The program consists of two phases. First, it determines whether the coaching message contains any violations, using five predicates for the dimensions that require semantic judgment and two deterministic functions that verify the word-count and sentence-count rubrics. The result is stored in \texttt{detected\_violations}. In the second phase, it compares this result with the evaluator's verdict and reasoning, verifying that either both are clean or that every violation cited by the evaluator is supported by a detected violation. This comparison uses two predicates that audit the evaluator's evidence, operating on the message and each piece of evidence independently. The resulting decision tree terminates in one of three outcomes: \textit{Pass}, \textit{Fail}, or \textit{Bug}.

\section{Dataset Details}
\label{sct:app_dataset_details}
This section complements the dataset descriptions in Section~\ref{sct:datasets} with the source of each dataset, the split sizes, and test label distribution.

The Coaching dataset was constructed by manually adapting real-world cases of a recommendation system into samples, covering a diverse range of scenarios that target different parts of the policy definition.
The task has three output classes with labels Pass, Fail and Bug. The training and test splits contain 103 and 71 samples. The test set comprises 29 Bug, 24 Fail and 18 Pass samples.

The Escalation dataset is based on the DynaGuard benchmark~\cite{hoover2025dynaguard}, which contains multi-turn dialogues annotated against custom policies. In the original policy, the agent escalates by returning a fixed string once repetition is detected. We generalized the definition of escalation to any reply connecting the user with a human agent, sharpened the notion of intent-based repetition, and extended the dataset with additional challenging human-annotated samples.
The training and test splits contain 85 and 286 samples. The test samples are divided into 147 True and 139 False. The True samples comprise 74 with three or more repeated intents followed by escalation and 73 with fewer than three repeated intents and no escalation. The False samples comprise 69 with three or more repeated intents but no escalation and 70 with escalation but fewer than three repeated intents.

The Plan dataset is built on traces of a research agent solving tasks from the GAIA benchmark \citep{mialon2024gaia}. Each trace is labeled Pass or Fail, according to whether the agent's plan is valid. In the original dataset, invalid plans fail in a single way, by omitting the \texttt{<end\_plan>} tag. We added authored samples that violate the other two conditions as well, referring to tools or collaborators absent from the trace and specifying explicit tool calls, as well as boundary valid samples that challenge each of the conditions. The training and test splits contain 82 and 81 samples, the latter divided into 41 Fail and 40 Pass.

The Tile-match dataset was constructed in a similar way to Coaching. Unlike the other tasks, it is a multi-label task with a sample-specific label space: each sample comes with its own list of candidate tiles, of varying length. The training and test splits contain 39 and 124 samples. Of the test samples, 45 refer to several tiles, 24 to a single tile, and 55 to none, designed as hard negatives.

\section{Cost Analysis}
\label{sct:app_cost_analysis}
Table~\ref{tab:cost} compares the inference cost of the two configurations studied in Section~\ref{sct:transfer}: a decomposition run by an inexpensive executor and a frontier reasoning model applied to the undecomposed task. For the decomposition, we assume pricing of \$0.15 per 1M input tokens, comparable to the list price of DeepSeek-V4.1-Flash; since decomposition runs without reasoning, we treat its output cost as negligible. We compare to gemini-3.1-pro with dynamic thinking, the cheaper of the two frontier models by list price (\$2 and \$12 per 1M input and output tokens vs. \$5 and \$30 for gpt-5.5). Although the decomposition makes several predicate calls per sample, it costs 9 to 47 times less.

\begin{table}[!t]
\centering
\caption{Comparison of inference costs of decompositions run by an inexpensive executor and of gemini-3.1-pro on the undecomposed tasks. Ratios computed before rounding.}
\label{tab:cost}

\begin{tabular}{lccc}
\addlinespace[0.2em]
\toprule
 & \multicolumn{2}{c}{\textbf{Cost per 1000 samples (\$)}} & \\
\cmidrule(lr){2-3}
\textbf{Task} & \textbf{Decomposed} & \textbf{gemini-3.1-pro} & \textbf{Ratio ($\times$)} \\
\midrule
Coaching   & 0.80 & 6.88 &  9 \\
Escalation & 0.23 & 10.71 & 46 \\
Plan       & 1.49 & 21.99 & 15 \\
Tile-match & 1.08 & 50.79 & 47 \\
\bottomrule
\end{tabular}
\end{table}

\section{Detailed Results by Executor Model}
\label{sct:app_detailed_results}
\begin{table}[H]
\centering
\caption{Test accuracy on four classification tasks under three executor models, configured to run with no reasoning. For the Coaching, Escalation and Plan tasks the accuracy is computed by the percentage of correct predictions. For the Tile-match task, the accuracy is computed as the mean-jaccard index between the correct labels and the predicted labels, over all samples. All reported numbers are based on 9 measurements, averaged over three independent runs of the optimization process and three repetitions of the test evaluation. The final numbers are rounded to two decimal places.}
\label{tab:experiments}
\begin{tabular}{l l cccc}
\addlinespace[0.5em]
\toprule
\textbf{Model} & \textbf{Method} & \textbf{Coaching} & \textbf{Escalation} & \textbf{Plan} & \textbf{Tile-match} \\
\midrule
\multirow{10}{*}{\textbf{gpt-5.4-nano}}
  & \textit{Direct prompting} & 0.60 & 0.56 & 0.59 & 0.38 \\
  \cmidrule(l){2-6}
  & \multicolumn{5}{l}{\textit{Prompt optimization}} \\
  \addlinespace[0.2em]
  & GEPA         & 0.58 & 0.61 & 0.66 & 0.55 \\
  \addlinespace[0.15em]
  & MIPROv2      & 0.69 & 0.57 & 0.58 & 0.56 \\
  \cmidrule(l){2-6}
  & \multicolumn{5}{l}{\textit{Task decomposition}} \\
  \addlinespace[0.2em]
  & CheckEval   & N/A & 0.58 & 0.54 & N/A \\
  \addlinespace[0.15em]
  & RLIE        & N/A & 0.58 & 0.73 & N/A \\
  \addlinespace[0.15em]
  & SCOPE (ours) & 0.85 & 0.82 & 0.87 & 0.78 \\
  \cmidrule(l){2-6}
  & \multicolumn{5}{l}{\textit{Workflow generation}} \\
  \addlinespace[0.2em]
  & AFlow        & 0.83   & 0.86   & 0.75   & 0.68   \\
\midrule
\multirow{10}{*}{\textbf{gpt-5.4-mini}}
  & \textit{Direct prompting} & 0.67 & 0.66 & 0.65 & 0.44 \\
  \cmidrule(l){2-6}
  & \multicolumn{5}{l}{\textit{Prompt optimization}} \\
  \addlinespace[0.2em]
  & GEPA         & 0.76 & 0.65 & 0.81 & 0.62 \\
  \addlinespace[0.15em]
  & MIPROv2      & 0.78 & 0.64 & 0.80 & 0.74 \\
  \cmidrule(l){2-6}
  & \multicolumn{5}{l}{\textit{Task decomposition}} \\
  \addlinespace[0.2em]
  & CheckEval   & N/A & 0.66 & 0.82 & N/A \\
  \addlinespace[0.15em]
  & RLIE        & N/A   & 0.59   & 0.89   & N/A   \\
  \addlinespace[0.15em]
  & SCOPE (ours) & 0.86 & 0.89 & 0.93 & 0.76 \\
  \cmidrule(l){2-6}
  & \multicolumn{5}{l}{\textit{Workflow generation}} \\
  \addlinespace[0.2em]
  & AFlow        & 0.81   & 0.92   & 0.81   & 0.70   \\
\midrule
\multirow{10}{*}{\textbf{gemini-3.5-flash}}
  & \textit{Direct prompting} & 0.87 & 0.69 & 0.85 & 0.49 \\
  \cmidrule(l){2-6}
  & \multicolumn{5}{l}{\textit{Prompt optimization}} \\
  \addlinespace[0.2em]
  & GEPA         & 0.87 & 0.86 & 0.88 & 0.89 \\
  \addlinespace[0.15em]
  & MIPROv2      & 0.93 & 0.72 & 0.89 & 0.89 \\
  \cmidrule(l){2-6}
  & \multicolumn{5}{l}{\textit{Task decomposition}} \\
  \addlinespace[0.2em]
  & CheckEval   & N/A & 0.70 & 0.93 & N/A \\
  \addlinespace[0.15em]
  & RLIE        & N/A & 0.92 & 0.93 & N/A \\
  \addlinespace[0.15em]
  & SCOPE (ours) & 0.93 & 0.95 & 0.93 & 0.89 \\
  \cmidrule(l){2-6}
  & \multicolumn{5}{l}{\textit{Workflow generation}} \\
  \addlinespace[0.2em]
  & AFlow        & 0.88   & 0.87   & 0.83   & 0.86   \\
\bottomrule
\end{tabular}
\end{table}

\clearpage

\section{Adversarial examples}
We present two adversarial examples, one from each level of the adversary: a systemic sample, which exposes a gap in how the composition program combines predicates, and a local sample, which exposes a single predicate misjudging an input within its own scope. Both examples target intermediate decompositions produced during the search.

\subsection{Systemic adversarial examples}
\label{sct:app_adversarial_systemic}
\begin{artifactbox}{Task definition}
A fitness and wellness recommendation system generates short coaching
messages that nudge users toward healthier habits. An automated critic then
judges each message against a style-and-safety rubric, emitting a verdict and
cited reasoning. Your job is to audit that critic: given a message and the
critic's evaluation of it, decide whether the critic's judgment is correct.

\medskip
Your task is to classify a given sample into exactly one of three labels:
\texttt{'Pass'}, \texttt{'Fail'}, or \texttt{'Bug'}. A \texttt{'sample'} consists
of a fitness-coaching message together with the underlying evaluator's verdict
and cited reasoning (e.g., word count, fired rule). You must always select the
best-fitting label.

\medskip
Seven Rubric Dimensions (for reference):

\begin{itemize}
    \item word\_count: under 18 words
    \item sentence\_count: a single sentence
    \item actionable: recommends a concrete and relevant exercise related action the user can carry out
    \item non\_judgmental: don't use judgmental language
    \item no\_gamification: avoid messages that include gamification
    \item safety\_first: does not require the user to do actions that are unsafe for their body or current state (e.g., push through sharp pain)
    \item system\_grounded: if the \texttt{system\_state} field is non-empty then the \texttt{coaching\_message} must address the advice category or the user's weakest dimension - any unrelated topic violates this.
\end{itemize}

\medskip
Label Definitions \& Boundaries:

\vspace{2mm}
\textbf{Pass:}

Positive Definition: The underlying coaching message adheres to ALL seven rubric dimensions AND the evaluator correctly accepted it.

Boundary Rules: Only label \texttt{'Pass'} when both the message is clean and the evaluator agrees.

\medskip
\textbf{Fail:}

Positive Definition: The coaching message violates at least one of
the seven rubric dimensions AND the evaluator correctly rejected it, citing an applicable reason. Any word count the evaluator cites must be computed on the candidate message alone.

Boundary Rules: Only label \texttt{'Fail'} when the violation is real and the evaluator's cited reason matches it.

\medskip
\textbf{Bug:}

Positive Definition: The evaluator's output is wrong in any of the
following ways:
\begin{itemize}
    \item False positive: a clean message was rejected.
    \item False negative: a violating message was accepted.
    \item Right verdict, wrong reason: the message truly violates some
    dimension, but the reason cited by the evaluator does not apply.
    \item Inconsistent verdict/reasoning: if \texttt{eval\_verdict} is
    \texttt{'Fail'} then \texttt{eval\_reasoning} must cite at least one
    reason; a \texttt{'Pass'} that nonetheless cites a violation is a
    \texttt{'Bug'}.
\end{itemize}

Boundary Rules: Any of the above sub-cases makes the sample a
\texttt{'Bug'}. A Fail verdict that happens to land on the right outcome via incorrect reasoning is still a Bug, not a Fail.

\bigskip
Sample Layout:

\vspace{0.8mm}
Each sample is a series of XML-tagged blocks of the form \texttt{<field\_name>value</field\_name>}, one per field. Fields may be omitted when not present.

\begin{itemize}
    \item \texttt{system\_state}: optional user profile / state / advice
    category / intensity metadata (already separated from the coaching message).
    \item \texttt{coaching\_message}: the candidate message text only (this is
    what the rubric dimensions are checked against).
    \item \texttt{eval\_verdict}: \texttt{'Pass'} or \texttt{'Fail'}.
    \item \texttt{eval\_reasoning}: a JSON array listing each violated dimension
    the evaluator flagged, e.g.
    \texttt{[{"Dimension": "non\_judgmental", "Evidence": "you always"}, ...]}.
    Each entry has a Dimension (one of the seven rubric dimensions) and
    Evidence --- either a specific snippet from the \texttt{coaching\_message}
    or a short free-text explanation of why that dimension is violated. Its
    value is empty when \texttt{eval\_verdict} is \texttt{'Pass'}.
\end{itemize}

\medskip
Decision Procedure:

\begin{enumerate}
    \item Check the \texttt{coaching\_message} against each of the seven
    rubric dimensions to determine whether it should Pass or Fail.
    \item Compare the \texttt{eval\_verdict} to step 1's conclusion.
    \item If verdicts disagree, the label is \texttt{'Bug'} (false positive or
    false negative).
    \item If verdicts agree and the verdict is Pass, the label is
    \texttt{'Pass'}.
    \item If verdicts agree and the verdict is Fail, verify that every entry in
    \texttt{eval\_reasoning} names a Dimension that is genuinely violated by
    the \texttt{coaching\_message} AND its Evidence is consistent with that
    violation. If that check fails, the label is \texttt{'Bug'}; otherwise
    \texttt{'Fail'}.
    \item Output only the chosen label.
\end{enumerate}

\medskip
Examples:

\vspace{1.5mm}
Pass Positive Example (all dimensions met):
\begin{lstlisting}[style=xmlsample]
<coaching_message>Gently ease into the stretch to maintain a comfortable range.</coaching_message>
<eval_verdict>Pass</eval_verdict>
<eval_reasoning></eval_reasoning>
\end{lstlisting}

\vspace{2mm}
Fail Positive Example (real violation of non\_judgmental with correct evidence):
\begin{lstlisting}[style=xmlsample]
<coaching_message>You always miss the warm up.</coaching_message>
<eval_verdict>Fail</eval_verdict>
<eval_reasoning>[{"Dimension": "non_judgmental", "Evidence": "you always"}]</eval_reasoning>
\end{lstlisting}

\vspace{2mm}
Bug -- False Positive Example (the cited Evidence is not present in the message nor implied by it):
\begin{lstlisting}[style=xmlsample]
<coaching_message>Keep your focus by setting your playlist before you begin the session.</coaching_message>
<eval_verdict>Fail</eval_verdict>
<eval_reasoning>[{"Dimension": "non_judgmental", "Evidence": "stop checking your phone during sets"}]</eval_reasoning>
\end{lstlisting}

\vspace{2mm}
Bug -- Right Verdict, Wrong Reason Example (the cited Dimension and Evidence are wrong, the actual violation is no\_gamification):
\begin{lstlisting}[style=xmlsample]
<coaching_message>Ease into your pace to hold onto your top 10% ranking this week.</coaching_message>
<eval_verdict>Fail</eval_verdict>
<eval_reasoning>[{"Dimension": "non_judgmental", "Evidence": "you always"}]</eval_reasoning>
\end{lstlisting}

\medskip
Output Guidance: Return only the label (one of \texttt{['Pass', 'Fail', 'Bug']}). Do not output explanations, reasoning, or additional formatting.
\end{artifactbox}

\begin{artifactbox}{Decomposition}
\begin{innerbox}{Predicate \texttt{p\_message\_compliant}}
You are auditing a fitness coaching message for compliance with five rubric dimensions. The message may be accompanied by a system\_state field describing the user's profile, current state, advice category, or weakest fitness dimension.

\medskip

Do NOT evaluate word count or sentence count — those are checked separately.

\medskip

Decide whether the coaching message complies with ALL FIVE of the following dimensions. Answer "yes" if it complies with all five, or "no" if it violates at least one.

\medskip

-- DIMENSION 1: actionable ...

-- DIMENSION 2: non\_judgmental ...

-- DIMENSION 3: no\_gamification ...

-- DIMENSION 4: safety\_first ...

-- DIMENSION 5: system\_grounded ...

\medskip

ANSWER: "yes" if the coaching message complies with ALL FIVE dimensions above, "no" if it violates at least one.
\end{innerbox}

\begin{innerbox}{Predicate \texttt{p\_reasoning\_entry\_valid}}
You are auditing a single reasoning entry that an automated evaluator cited when rejecting a fitness coaching message. Your job is to determine whether this specific reasoning entry is valid.
You will receive: ... coaching\_message, system\_state, cited dimension, cited evidence...

\medskip

A reasoning entry is VALID ("yes") if and only if BOTH of the following hold:
\begin{enumerate}
    \item The cited Dimension is genuinely violated by the coaching message.
    \item The cited Evidence accurately reflects the violation --- either as a
    verbatim snippet from the coaching message OR as a logically sound
    free-text explanation of why that dimension is violated. The evidence does
    NOT need to quote the message verbatim; a paraphrased or approximate
    description is acceptable as long as it accurately characterizes the
    actual violation.
\end{enumerate}

A reasoning entry is INVALID ("no") if:
\begin{itemize}
    \item The cited Dimension is NOT actually violated by the coaching message
    (the violation does not exist or is fabricated), OR
    \item The cited Evidence does not accurately reflect the actual violation
    (e.g., it quotes text not present in the message, describes behavior the
    message does not exhibit, or identifies the wrong infraction).
\end{itemize}

IMPORTANT: You are evaluating only THIS ONE entry. Whether the message has other violations that were not cited is irrelevant, judge only whether this specific citation is accurate.

RUBRIC DIMENSIONS you may encounter: actionable, non\_judgmental, no\_gamification, safety\_first, system\_grounded.

\medskip
ANSWER: ``yes'' if this reasoning entry is valid, ``no'' if it is invalid.
\end{innerbox}
\end{artifactbox}

\begin{artifactbox}{Combination code}
\begin{lstlisting}[style=pythonstyle]
def evaluate(predict, sample):
    msg = get_coaching_message(sample)
    verdict = get_eval_verdict(sample)
    entries = get_eval_reasoning_entries(sample)
    system_state = get_system_state(sample)

    # === Step 1: Deterministic dimension checks ===
    # word_count: must be strictly under 18 words (17 or fewer);
    # exactly 18 is a violation
    wc = word_count(msg)
    wc_violation = wc >= 18

    # sentence_count: must be a single sentence
    sent_violation = has_multiple_sentences(msg)
    det_violation = wc_violation or sent_violation

    # === Step 2: Subjective dimension compliance ===
    if system_state:
        compliance_ctx = ("<system_state>" + system_state +
                          "</system_state>\n"
                          "<coaching_message>" + msg +
                          "</coaching_message>")
    else:
        compliance_ctx = "<coaching_message>" + msg + "</coaching_message>"

    subj_ok = predict('p_message_compliant', compliance_ctx)

    # === Step 3: Overall message compliance determination ===
    message_should_fail = det_violation or (subj_ok == "no")
    evaluator_says_fail = (verdict == "Fail")

    # === Step 4: Verdict mismatch -> Bug ===
    if message_should_fail != evaluator_says_fail:
        return "Bug"

    # === Step 5: Both verdicts agree on Pass ===
    if not message_should_fail and not evaluator_says_fail:
        if entries:
            return "Bug"
        return "Pass"

    # === Step 6: Both verdicts agree on Fail ===
    if not entries:
        return "Bug"

    for entry in entries:
        if not isinstance(entry, dict):
            return "Bug"

        dim = entry.get('Dimension', '')
        evidence = entry.get('Evidence', '')

        if dim == 'word_count':
            if not wc_violation:
                return "Bug"

        elif dim == 'sentence_count':
            if not sent_violation:
                return "Bug"

        else:
            # Subjective dimension -- LLM validates this entry
            if system_state and dim == 'system_grounded':
                entry_ctx = ("system_state: " + system_state + "\n"
                             "coaching_message: " + msg + "\n"
                             "Cited dimension: " + dim + "\n"
                             "Cited evidence: " + evidence)
            else:
                entry_ctx = ("coaching_message: " + msg + "\n"
                             "Cited dimension: " + dim + "\n"
                             "Cited evidence: " + evidence)

            if predict('p_reasoning_entry_valid', entry_ctx) == "no":
                return "Bug"

    return "Fail"

def has_multiple_sentences(text):
    import re
    if not text:
        return False
    if ';' in text:
        return True
    if re.search(r'[.!?]\s+[A-Z]', text):
        return True
    return False

def word_count(text):
    if not text:
        return 0
    return len(text.split())
\end{lstlisting}
\end{artifactbox}

\begin{artifactbox}{Adversarial example: Fabricated evidence on a deterministically checked dimension}
\begin{innerbox}
\textbf{Sample} (gold: \texttt{Bug}; decomposition output: \texttt{Fail}):
\begin{lstlisting}[style=xmlsample]
<coaching_message>Raise your arms slowly overhead. Lower them back down with control.</coaching_message>
<eval_verdict>Fail</eval_verdict>
<eval_reasoning>[{"Dimension":"sentence_count", "Evidence":"contains three separate statements joined by semicolons"}]</eval_reasoning>
\end{lstlisting}

\medskip

\textbf{The gap.} The combination code detects the sentence-count violation deterministically (\texttt{has\_multiple\_sentences}). But unlike the subjective dimensions, whose cited evidence is checked by the \texttt{p\_reasoning\_entry\_valid} predicate, the \texttt{sentence\_count} branch only confirms that a violation \emph{exists}; it never verifies that the cited evidence matches the message. Here the message genuinely violates \texttt{sentence\_count} (two sentences split by a period), yet the evaluator's evidence is fabricated (three statements joined by semicolons). The rule labels a right-verdict / wrong-evidence case as \texttt{Bug}, but the combination code accepts the entry and outputs \texttt{Fail}.
\end{innerbox}
\end{artifactbox}

\subsection{Predicate-local adversarial examples}
\label{sct:app_adversarial_local}
\begin{artifactbox}{Task definition}
You are given a complete user-agent conversation transcript (alternating <User> and <Agent> turns, ending with the agent's reply). Output exactly one label, True or False, judging whether the agent handled escalation correctly.

\medskip

Escalation is warranted if the user re-raises the same unresolved intent at least three times - verbatim, rephrased, or as renewed dissatisfaction that it still isn't resolved (e.g. "that didn't work", "you're not helping"). A turn counts unless it moves the request forward - a new constraint, a new distinct question, or new information that changes the situation; simply restating the unmet intent, including the agent's failure to address it, still counts.

\medskip

A conversation is considered escalated when the agent's final reply connects the user with a human agent; any other final reply (answering, declining, asking something back, etc.) is not an escalation.

\medskip

Assign the label by comparing what the agent actually did in its final reply against what was warranted:
\begin{itemize}
    \item \textbf{True}: the agent acted correctly - an escalation was both warranted and performed, or an escalation was not warranted and was not performed by the agent.
    \item \textbf{False}: the agent acted incorrectly - escalation is warranted but was not performed by the agent, or no escalation is needed but the agent performed one.
\end{itemize}
\end{artifactbox}

\begin{artifactbox}{Decomposition}
\begin{innerbox}{Predicate \texttt{p\_warranted}}
You are given a complete user-agent conversation transcript (alternating User and Agent turns). Your task is ONLY to determine whether the user raised the same single unresolved intent at least three times.

A user turn COUNTS toward the threshold if the user:
\begin{itemize}
\item Repeats the same request verbatim
\item Rephrases the same unmet demand in different words
\item Expresses renewed dissatisfaction that the issue remains unresolved
\item Responds to the agent's clarifying question by declining to elaborate and simply reiterating the original request — this is a repetition, not new information.
\item Reports that the agent's suggested solution does not apply to them or has not worked - foreclosing one proposed path does not change what the user fundamentally needs; the underlying intent is unchanged and the turn counts
\item Adds motivation, urgency, or personal context to an otherwise unchanged request
\end{itemize}

A user turn does NOT count if it:
\begin{itemize}
\item Introduces a new constraint or requirement that changes what specifically is being requested
\item Asks a new, distinct question or shifts to a different topic or sub-topic
\item Provides new information that materially changes the situation, meaning information that changes what the user fundamentally needs or what the underlying problem is
\end{itemize}

Important counting rules:
\begin{itemize}
\item Count per distinct intent separately.
\item Whether the agent produced any response is NOT the criterion for 'resolved.'
\item Do not rely on the user's own claim about how many times they've asked.
\end{itemize}

Answer 'yes' if at least one single unresolved intent was raised by the user at least three times. Answer 'no' if no single unresolved intent reached that threshold.
\end{innerbox}

\begin{innerbox}{Predicate \texttt{p\_escalated}}
You are given the text of an agent's reply in a customer-service conversation. Your task is ONLY to determine whether this reply connects the user with a human agent. Answer 'yes' if the reply:

\begin{itemize}
\item Explicitly transfers the user to a live human agent or operator
\item Creates or opens a support ticket that will be handled by a human specialist
\item Informs the user that a human agent will contact them
\item Provides a link that the agent explicitly characterizes as an immediate live-chat connection to a human agent, distinct from a general support portal, help-center URL, or phone number
\end{itemize}

Answer 'no' if the reply:
\begin{itemize}
\item Answers the user's question or provides information
\item Declines the request based on policy or inability
\item Asks the user a clarifying question
\item Apologizes or expresses sympathy without connecting to a human
\item Provides a phone number, email address, or general support URL for the user to contact support themselves
\end{itemize}
\end{innerbox}
\end{artifactbox}

\begin{artifactbox}{Combination code}
\begin{lstlisting}[style=pythonstyle]
def evaluate(predict, sample):
    # Assess whether escalation was warranted (full conversation needed)
    warranted = predict('p_warranted', sample)

    # Assess whether the agent's final reply escalated (only final turn needed)
    last_agent = get_last_agent_turn(sample)
    escalated = predict('p_escalated', last_agent)

    # True iff agent action matches warrant status
    # (warranted + escalated) OR (not warranted + not escalated)
    if (warranted == 'yes') == (escalated == 'yes'):
        return 'True'
    return 'False'
\end{lstlisting}
\end{artifactbox}

\begin{artifactbox}{Adversarial example for predicate p\_warranted: refined user intent}
\textbf{Sample} (gold: \texttt{True}; decomposition output: \texttt{False}):

\medskip

\begin{lstlisting}[style=xmlsample]
User: Hi, I ordered a blender two weeks ago and it still hasn't arrived. I'd like a refund, please.

Agent: I'm sorry to hear that. Let me help - have you checked the tracking link in your order confirmation email to see the latest status?

User: Okay, I just checked the tracking and now it says the package was delivered three days ago and signed for by someone at my address. I never actually received it, so I still want a refund.

Agent: Thanks for checking. When a package shows as delivered and signed for, the carrier records it as completed, so this becomes a disputed-delivery case rather than a lost package. I'd suggest first checking with anyone else at your address or neighbors who may have accepted it.

User: I've asked everyone and no one has it. I still want my refund for the blender I never received.

Agent: Understood. I've logged your report about the missing blender. Please allow 3 to 5 business days while we review the delivery records with the carrier. Is there anything else I can help you with?
\end{lstlisting}

\medskip

\textbf{The gap.} The rule excludes a turn that supplies 'new information that changes the situation'. The second user turn materially changes the situation from lost parcel to disputed delivery, so that turn does not count for repetition. Thus, escalation is not warranted and the agent acted correctly. On the decomposition side, the predicate p\_warranted narrows 'changes the situation' down to 'changes what the user fundamentally needs / the underlying problem', interpreting the second turn as a repetition and the overall classification as False.
\end{artifactbox}
\clearpage
\section{Task decomposition prompts}
\label{sct:app_prompts}
\begin{artifactbox}{Analyzer}
\noindent\textbf{\texttt{SYSTEM PROMPT}}
\begin{lstlisting}[style=prompt]
Your job is to analyze the failures of a decomposed classification rule on the user's real labeled data — diagnose what is going wrong (across sub-predicates, helpers, and combination_code) and recommend corrective changes for the decomposer to weigh.

You will receive:
- The original classification rule.
- The current decomposition: classifier sub-predicates (each PURE — just a question + its label set; it carries no input slicing), reusable deterministic helpers (each a Python `def <name>(...)` over `sample`), and the Python combination_code `def evaluate(predict, sample)` that slices the sample (via helpers), calls `predict(name, text)` to classify slices, and combines the results into the final label.
- SAMPLE VERDICTS covering EVERY training sample of this iteration, SPLIT into two sections: FAILURES (the misclassified samples — what to fix) and CORRECT (the contrast set). Per sample it gives: `sample_index`, its `calls` (the ORDERED list of predict() calls the combination_code made — each shown as predicate_name→label; a predicate may appear zero, one, or many times), the `combined_label` the decomposition produced, and the `expected_final_label` (ground truth). This is the full picture of what every predict call returned and where the decomposition was right or wrong — the CORRECT rows are your contrast set: before committing to a move, identify the CORRECT rows CLOSE to the failure you're fixing (related `reasoning`) and FETCH them too, then counter-hypothesize the move against them — confirm it would not flip any of them (regression). The verdicts do NOT include the call INPUT text.
- The Read and Bash tools, plus a CSV file (path in <SAMPLE_TEXT_ACCESS>) holding the full text of every sample and every predict call. To inspect a sample's actual input, FETCH it from the CSV — read only the samples you need to understand a failure; you do not need to (and should not) read them all.

HOW TO FETCH SAMPLES' TEXT: use the Bash tool with pandas (bare `python3` lacks pandas — use `uv run python`). Collect ALL the sample_indexes you want — the failures you are diagnosing PLUS their close-correct contrast rows (see below) — and fetch them in ONE call by passing the whole LIST, so the file is read once:
  uv run python -c "import pandas as pd; d=pd.read_csv('<PATH>'); r=d[(d.iteration==<ITER>)&(d.sample_index.isin([<K1>,<K2>,...]))]; print(r[['sample_index','call_index','predicate_name','input_text','label','original_sample']].to_string())"
The CSV has one row per (sample, predict-call): `original_sample` is the full input block; `input_text` is the exact slice that predict() call classified; `predicate_name`/`label` are that call's predicate and returned label; `call_index` orders the calls within the sample. Batch the ids — do NOT run one command per id. The CSV holds ONLY sample text + per-call slices; `reasoning` lives in the verdict table above, not the CSV.

CONTEXT / SLICING is a common failure cause: the combination_code may feed a predicate the wrong slice (too much irrelevant text, or a slice that dropped the region the sub-question needs). When you suspect it, fetch that sample's `original_sample` and the offending call's `input_text` and compare what the classifier saw against the region the sub-question needs. The slicing lives in combination_code (and the helper it calls) — so recommend 'rewrite_combination' (or 'rewrite_helper' / 'add_helper' for the slicing function), naming the region in the rationale. To change what a predicate DECIDES (not what it reads), use 'reword'.

The verdict columns per sample are: `sample_index`; `calls` (ordered predicate_name→label for each predict() the combination_code made); `combined_label` (the final label combination_code produced — 'COMBINE_ERROR' means the combination raised on this sample, a bug in the code/helpers → rewrite_combination); and `expected_final_label` (ground truth). Which section a row is in (FAILURES vs CORRECT) tells you whether it was misclassified. For DYNAMIC-label tasks a `candidates` column also appears: this sample's allowed final labels (the per-sample label space both `combined_label` and `expected_final_label` live in). Use it to spot combiners that return a label OUTSIDE the candidates, or fail to pick the right candidate that was available. For MULTI-LABEL tasks `combined_label` and `expected_final_label` are comma-joined SETS (a subset of the candidates; empty string = empty set); rows also show `missed` (in gold, not predicted) and `spurious` (predicted, not in gold) plus `jaccard` (set overlap, 1.0 = exact). FAILURES (jaccard<1.0) are sorted worst-first. `missed` means the combiner UNDER-selected (a membership test too strict, or a candidate never checked); `spurious` means it OVER-selected (a membership test too loose).

For each FAILED sample, follow this reasoning chain:
1. Read the `calls` (predict results) and mentally execute combination_code on them; confirm it produces the combined_label. If executing it does NOT reproduce combined_label, or if its logic (or a helper it calls) does not faithfully encode the original rule, the bug is in the combination itself — recommend rewriting it (target 'combination_code', move 'rewrite_combination', or 'helper:<name>' → 'rewrite_helper' if a slicing/value helper is wrong) and stop diagnosing sub-predicates for this sample.
2. Working backwards from expected_final_label through combination_code, identify which predict() result(s) would have had to differ to produce the expected label. This tells you which call(s) — hence which predicate on which slice — were wrong.
3. For each implicated call, FETCH the sample's rows from the CSV, look at the `input_text` that call was fed, and judge against the predicate's description whether its label actually looks wrong, and characterize WHY (what about this input tripped the weak classifier, or whether it was fed the wrong slice). If it was fed the wrong slice, the fix is in combination_code / the slicing helper, not the predicate.

If failures across samples consistently point at combination_code or a helper (not specific predicate judgements), recommend rewriting it (target 'combination_code' / 'helper:<name>') describing what is wrong.

Aggregate the per-sample reasoning across all failed samples into TWO outputs:
1. `analysis_text` — a concise narrative: which components are implicated, the failure patterns across samples, and why, all grounded in the ORIGINAL RULE. Reason about every classifier predicate (including the ones that work), but state patterns generally; do not quote samples verbatim.
2. `recommendations` — structured corrective moves, ONE per target you want changed. Each has a `target`, a `move`, and a `rationale` (the root-cause fix). Targets and moves:
- target a predicate ('p0', 'p1', …) with 'reword' (its judgement is worded ambiguously/incompletely — clarify/extend wording) or 'split' (OVERLOADED with several independent conditions — break into single-concept sub-questions);
- target 'combination_code' with 'rewrite_combination' when its logic — including which slice it feeds a predicate — is wrong;
- target 'helper:<name>' with 'rewrite_helper' when a slicing/value helper is wrong;
- add a 'new_predicate:<name>' / 'new_helper:<name>' (move 'add_predicate' / 'add_helper') when the rule depends on a condition/value nothing currently captures; or 'remove_predicate' / 'remove_helper' for a redundant one.
Pick the move that targets the ROOT cause: don't say 'split' for a wording problem, or 'reword' for a slicing or structural problem (that's rewrite_combination / rewrite_helper). You need NOT emit a 'keep' for every working predicate — only list targets you want changed. When there ARE failed samples, include at least one corrective recommendation.

RULE-FIRST DIAGNOSIS: the samples you see (failed or correct) are EVIDENCE of how the current predicate behaves — they are not the thing the predicate should be patched to fit. Every diagnosis and recommendation must be traced back to the ORIGINAL RULE: which clause, concept, definition, exception, or piece of scope in the rule the sub-predicate failed to capture or misapplied. State the failure mode in terms of what the rule says (or implies) that the predicate's current wording does not encode. Do NOT recommend additions whose only justification is a specific word, phrase, value, or example observed in any sample; if a pattern appears across samples, identify the part of the rule that already covers it and recommend that the predicate align with the rule there. If no part of the rule supports a proposed addition, do not recommend it — the rule is the ground truth, not the samples. Cite specific samples only to illustrate the failure mode; the recommendation itself must read as guidance grounded in the rule, not as a sample-pattern patch.

Do NOT prescribe a specific rewritten decomposition. Your job is to diagnose and recommend; the decomposer decides and designs the next decomposition.
\end{lstlisting}

\bigskip
\noindent\textbf{\texttt{--- USER PROMPT ---}}
\medskip
\begin{lstlisting}[style=prompt]
<ORIGINAL_RULE>
{task}
</ORIGINAL_RULE>

<POSSIBLE_FINAL_LABELS>{labels}</POSSIBLE_FINAL_LABELS>

{multi_note}
<CURRENT_SUB_PREDICATES>
{predicate_descriptions}
</CURRENT_SUB_PREDICATES>

<CURRENT_HELPERS>
{helper_descriptions}
</CURRENT_HELPERS>

<COMBINATION_CODE>
{combination_code}
</COMBINATION_CODE>

<SAMPLE_VERDICTS>
Overall train accuracy this iteration: {total_accuracy}. Use it to size the fix: high accuracy implies few, specific failures — prefer a targeted change and check it would not flip any CORRECT row (regression); low accuracy implies a systemic problem that may warrant a structural change.
Verdict rows are SPLIT into FAILURES (diagnose these) and the CORRECT contrast set (do not regress these). Columns per row: sample_index | calls | combined_label | expected_final_label | reasoning (the gold rationale for the label, when present — a primary fix hint). Use reasoning to find the CORRECT rows CLOSE to each failure. The call INPUT text is not here — fetch the failures AND their close-correct rows from the CSV in ONE batched call (see below).

FAILURES (the misclassified samples to fix):
{failed_verdicts}

CORRECT (contrast set — a proposed fix must not flip any of these):
{correct_verdicts}
</SAMPLE_VERDICTS>

<SAMPLE_TEXT_ACCESS>
Every sample's full text and every predict call is in this CSV (one row per sample * predict-call):
path: {csv_path}
filter: iteration == {iteration}
Fetch rows with the Bash tool (use `uv run python`, not bare python3) — pass the WHOLE list of sample_indexes you need (failures + their close-correct rows) so the file is read once:
uv run python -c "import pandas as pd; d=pd.read_csv('{csv_path}'); r=d[(d.iteration=={iteration})&(d.sample_index.isin([K1,K2,...]))]; print(r[['sample_index','call_index','predicate_name','input_text','label','original_sample']].to_string())"
Replace [K1,K2,...] with the sample_indexes you want and batch them — do not run one command per id. `original_sample` = the full input block; `input_text` = the exact slice that predict() call classified; `predicate_name`/`label` = that call's predicate and returned label; `call_index` orders the calls. The CSV has ONLY sample text + per-call slices — reasoning is in the verdict table, not here. Read only the samples you need.
</SAMPLE_TEXT_ACCESS>
\end{lstlisting}
\end{artifactbox}

\begin{artifactbox}{Decomposer}
\noindent\textbf{\texttt{SYSTEM PROMPT}}
\begin{lstlisting}[style=prompt]
Your job is to decompose a classification rule into sub-predicates so that a weak classifier can evaluate each sub-predicate independently, and deterministic code can combine their results to reproduce the original rule's decision.

The original rule is the ground truth — you must not change, relax, or reinterpret it.  The decomposition is only a structural aid: break one hard question into several simpler questions whose combined answer is identical to what the original rule would decide for every possible input.

You have THREE distinct kinds of building blocks; using the right one for the right subproblem is critical:

1. CLASSIFIER PREDICATES (`predicates`): the weak LLM picks one label from a small enumeration. Use these for subjective judgements. Every predicate's label set MUST be a FIXED, CLOSED enumeration you specify up front — typically "yes"/"no" or a small enum, THE SAME for every input. A weak classifier can only pick from a fixed set it is shown, so a predicate's labels must NEVER be per-sample, open-ended, or dynamic. When the final answer comes from a per-sample candidate set, don't put that set into a predicate's labels; instead have combination_code enumerate the candidates and decide among them using fixed-label predicate(s) — you choose the structure. A predicate is PURE: it carries NO input slicing. It does not decide what text it reads — the combination_code does, by calling `predict(name, text)` with whatever slice it wants. So the SAME predicate can be applied to many different slices.

2. HELPERS (`helpers`): reusable deterministic Python functions the combination_code calls. Each helper defines `def <name>(...)` (its function name must equal its `name`). A helper may take just `sample` or extra args (e.g. an index), and may return a VALUE (a count, a normalized form), a SLICE of the sample (the region a predicate should read), or a LIST of slices to loop over. All helpers are exec'd into the combination_code's namespace, so `evaluate` calls them directly by name. Helpers are deterministic — no LLM, no randomness/clock/IO. Use them for parsing/slicing the raw `sample` and for any value combination_code needs. If a judgement cannot be produced by deterministic Python alone, make it a classifier predicate instead.

3. COMBINATION CODE (`combination_code`): Python source defining `def evaluate(predict, sample) -> str`. The runner exec's the helper functions and this source into ONE namespace, then calls `evaluate(...)`.
     - `predict(predicate_name, text)` — classify `text` with predicate `predicate_name` (the unique `name` you gave that predicate; the names in combination_code MUST match the predicate `name` fields exactly) and return its label. Call it as many times as needed, including in a loop over slices; the same predicate may be called on different slices.
     - `sample` — the input-block TEXT (parse it, or use helpers).
     Helper functions are in scope — call them by name. Must return the final label string. Use combination_code for ALL deterministic logic — slicing (via helpers), looping, threshold comparisons, lookups, control flow. It MUST be deterministic so it behaves identically each run.

REPEATED SECTIONS (an option, not a rule): when the sample contains several same-format sections (e.g. repeated blocks/turns/items) and the analysis suggests a predicate is struggling because it sees all of them at once, consider calling that predicate in a LOOP — a helper returns the list of sections and combination_code calls `predict` once per section (feeding just the relevant section), then combines the per-section labels (e.g. any/all/count). Feed a section WITH surrounding sections when cross-section context is needed to judge it. Only reach for this when the evidence points to it; a single call on the whole sample is the default.

The `sample` text is raw — to operate on a specific part (a field, a turn, a tag's contents) parse it out with a helper (or inline).

Approach:
  1. Read the original rule and identify the distinct conditions it checks.
  2. For each condition, decide:
     - If the answer is a subjective judgement → classifier predicate (fed the right slice by combination_code).
     - If you need a value or a slice (a count, a fragment, the i-th section) → helper.
     - Stitch it together in combination_code: slice with helpers, call predicates via `predict`, combine deterministically.
  3. Write combination_code that exactly reconstructs the original rule.

PREFER an LLM classifier (not Python if/elif chains) when the test is semantic matching. String matching with hard-coded substrings is brittle; that is exactly what LLM classifiers are good at.

SELF-CONTAINED SUB-PREDICATES (CRITICAL): the weak classifier is shown ONLY a single sub-predicate's text — NOT the original rule. Therefore, each sub-predicate MUST be a COMPLETE, STANDALONE instruction that can be answered with ZERO external context: it states its own question, what each of its labels means, the relevant scope, and any needed examples/exceptions — on its own. When you author a sub-predicate, identify the portions of the original rule that define, constrain, or exemplify the concept that sub-predicate tests, and carry that information INTO the sub-predicate's own text (do NOT rely on the rule being present at classification time — it will not be). Preserve definitions, scope, examples, and exceptions. Do not introduce criteria that aren't derived from the original rule.

RULE-FIRST REFINEMENT: when applying the analyzer's recommendation, derive the refinement from the ORIGINAL RULE — not from the samples (failed or correct) the analyzer cited. If a recommendation proposes adding a specific word, phrase, value, or example whose only justification is that it appeared in a sample, do not add it; instead, identify the part of the rule that the analyzer's underlying observation maps to and rephrase the predicate so it captures that part of the rule generally. The decomposition must encode the rule's intent, not a list of patterns observed in the evaluation samples. The samples are a probe of the current decomposition's behavior; they are not training data and they are not the thing being fit.

DEFENSIVE code: parsing the sample text (or reading an extractor value) can raise on unexpected inputs (no marker/tag match, empty strings, non-numeric text). For numeric parsing, prefer a small helper such as `def _as_int(s):\n    try: return int(s)\n    except: return None` and gate downstream logic on `None`. When the marker/tag you expect is absent, default gracefully instead of indexing blindly.

Example — original rule: "If the user talks about politics, the agent should refuse politely to answer with max 10 words." (labels: yes/no; `sample` is the input-block TEXT, here of the form "Task input: <user message>\n\nModel response: <agent reply>"):
    Helpers:
      response: description="The agent's reply text.",
                deterministic_code=`def response(sample):\n    return sample.split("Model response:", 1)[-1].strip()`
      word_count: description="Number of words in the agent's reply.",
                  deterministic_code=`def word_count(sample):\n    return len(response(sample).split())`
    Classifier predicates (pure — combination_code feeds each the right slice):
      p0 = "Does the message discuss political topics" (yes/no)
      p1 = "Did the model refuse to answer" (yes/no)
      p2 = "Was the model's refusal polite" (yes/no)
    combination_code:
      def evaluate(predict, sample):
          # politics can appear anywhere → judge the whole exchange
          discusses_politics = predict("p0", sample)
          # refusal/politeness concern the reply → feed just that slice
          reply = response(sample)
          refused = predict("p1", reply)
          polite = predict("p2", reply)

          def handled_correctly():
              if not (refused == "yes" and polite == "yes"):
                  return False
              return word_count(sample) <= 10

          if discusses_politics == "yes" and not handled_correctly():
              return "no"
          return "yes"

RECOMMENDATIONS, NOT ORDERS — YOU DECIDE: the analysis reports where the COMBINED decomposition fails to reproduce the rule and RECOMMENDS changes (to predicates, helpers, and/or combination_code). YOU are the decision maker: weigh them against the overall picture and choose the change(s) that best move the decomposition toward reproducing the rule. You MAY override any single recommendation — including a 'keep' — when the overall picture justifies it. A failure means a condition the rule decides on is missing or mis-weighted, or a helper / combination_code is wrong — so a fix may implicate predicates, HELPERS, or combination_code. Act on real gaps, not contrived edge cases that would not occur in the real data.

AVAILABLE CHANGES: you are NOT limited to one move per predicate. You may reword a predicate, split it, ADD new predicates/helpers for an uncaptured condition, REWRITE helpers (including the slicing a predicate is fed), and REWRITE combination_code — whatever the evidence supports. To change what a predicate READS, change the combination_code (or the helper it slices with), not the predicate. Prefer focused, well-justified changes over sweeping rewrites, and preserve the predicate/helper outputs that the correct cases already rely on.
\end{lstlisting}

\bigskip
\noindent\textbf{\texttt{--- USER PROMPT ---}}
\medskip
\begin{lstlisting}[style=prompt]
<ORIGINAL_RULE>
{task}
</ORIGINAL_RULE>

<POSSIBLE_FINAL_LABELS>{labels}</POSSIBLE_FINAL_LABELS>

{dynamic_labels_section}<SAMPLE_INPUTS>
Real example values of the `sample` text your helpers receive (and that combination_code parses / feeds to `predict`) — use them to learn the input block's actual format and write parsing (regex/splits) that targets the right region. Do NOT tailor the decomposition to these specific cases; they are only a format reference (the rule is the ground truth).
{sample_examples}
</SAMPLE_INPUTS>

<ITERATION_HISTORY>
Previous decompositions and their accuracy on the train set:
{history}
</ITERATION_HISTORY>

<PREVIOUS_MOVES>
The set of changes YOU applied last turn (target: move — why). Don't repeat a change that didn't help; if a target's last move failed to improve things, try a different kind of move on it (or a different target). "(none …)" means this is the first improvement step.
{previous_actions}
</PREVIOUS_MOVES>

<FAILED_DIRECTIONS>
The search has backtracked to this decomposition because recent iterations did not improve on it. Each "Attempt" below is a COMPLETE move-set (across predicates, helpers, and combination_code) already tried FROM THIS decomposition that did NOT improve accuracy. Choose a SUBSTANTIALLY DIFFERENT direction — different targets and/or different kinds of moves; do not propose a minor variation of a failed attempt. "(none …)" means nothing has been tried from here yet, so proceed normally.
{failed_directions}
</FAILED_DIRECTIONS>

You are given the analyses below, each RECOMMENDING changes. They are inputs to YOUR decision, not orders — weigh them, reconcile any contradictions, and decide the best change(s). They localize the problem to (1) the decomposition on the user's real data, or (2) a structural gap vs the rule (which can implicate predicates, helpers, or combination_code).

<TRAIN_ANALYSIS>
TRAIN ANALYSIS — the COMBINED decomposition on the TRAIN set: a narrative diagnosis of where it produces the wrong final label, followed by structured recommendations (each: target → move — rationale, targeting a predicate / helper / combination_code, or a new predicate/helper to add). Treat these as recommendations you may follow or override.
{analysis}
</TRAIN_ANALYSIS>

Propose a new decomposition (helpers + classifier predicates + combination_code) that is logically equivalent to the original rule above. You are the DECIDER: using the analyses as recommendations, choose and apply the change(s) — across predicates, helpers, and combination_code — that best improve the decomposition. Follow a recommended per-predicate action when it is right, or override it (keep a predicate the analyses flagged, or change one they said to 'keep') when the overall picture warrants. The per-predicate moves below are your toolkit; apply each one as defined when you choose it:
- 'keep' — carry the predicate over verbatim (same wording and labels).
- 'reword' — refine ONLY that predicate's wording; keep its labels and combination_code unchanged. Use this to change what the predicate DECIDES — NOT what text it reads (that lives in combination_code).
- 'split' — replace that one predicate with the N finer single-concept predicates the analysis described, and rewrite combination_code to consume the new predicate outputs. A freshly-split child gets its initial wording now, but is NOT additionally reworded this turn — that can only happen in a later turn once it has been measured.
To change what a predicate is FED, do NOT touch the predicate — change combination_code (or the helper it slices with: add_helper / rewrite_helper). Change combination_code whenever your decision requires it — a split, an added or removed predicate, a helper change, or a direct combination fix.

Inside combination_code, `predict(name, text)` classifies `text` with predicate `name`; `sample` is the input-block text, described as: {sample_schema}; and every helper you declared is a function in scope (call it by name). Slice `sample` with helpers and feed the slices to `predict`.

Finally, populate `applied_moves`: list EVERY change you made this turn versus the base decomposition (one entry per changed predicate / helper / combination_code / added or removed predicate), each as {{target, move, rationale}}. This is recorded as the search direction so a later backtrack can avoid repeating it — so report what you actually changed, accurately.
\end{lstlisting}
\end{artifactbox}

\end{document}